%% file: main.tex
\documentclass[aap]{stylefile}

\RequirePackage{amsthm,amsmath,amsfonts,amssymb}
\RequirePackage[sort,numbers]{natbib}
\RequirePackage{mathtools}
\RequirePackage[shortlabels]{enumitem}
\RequirePackage{algorithm}
\RequirePackage{algorithmic}
\RequirePackage{tikz}
\RequirePackage[colorlinks,citecolor=blue,urlcolor=blue]{hyperref}
\RequirePackage{graphicx}
\usepackage[normalem]{ulem}
\usepackage{bbm}
\usepackage[normalem]{ulem}

\usepackage{multirow}
\usepackage{makecell, cellspace}
\usepackage{booktabs}
\usepackage[flushleft]{threeparttable}
\usepackage{booktabs} 
\usepackage{ragged2e}
\usepackage{bm}
\usepackage{multicol}
\usepackage{multirow}
\usepackage[makeroom]{cancel}
\usepackage{pifont}

\DeclareMathOperator*{\argmax}{arg\,max}

\usepackage[capitalize,noabbrev]{cleveref}
\startlocaldefs
\theoremstyle{plain}
\newtheorem{theorem}{Theorem}[section]

\newtheorem{lemma}{Lemma}[section]
\newtheorem{corollary}{Corollary}[theorem]
\newtheorem{proposition}{Proposition}[section]

\theoremstyle{definition}

\newtheorem{assumption}{Assumption}[section]

\theoremstyle{remark}

\newcommand{\E}{{\mathbb{E}}}

\endlocaldefs

\usepackage{hyperref}

\newcommand{\pii}{{\hat{\pi}}}
\newcommand{\CL}{{L_1}}
\newcommand{\CLL}{{L_2}}
\newcommand{\CLLL}{{L_3}}

\def\QEDclosed{\mbox{\rule[0pt]{1.3ex}{1.3ex}}} 
\def\QEDopen{{\setlength{\fboxsep}{0pt}\setlength{\fboxrule}{0.2pt}\fbox{\rule[0pt]{0pt}{1.3ex}\rule[0pt]{1.3ex}{0pt}}}}
\def\QED{\QEDclosed}

\begin{document}

\begin{frontmatter}
\title{Finite Sample Analysis of Two-Time-Scale
Natural Actor-Critic Algorithm}

\begin{aug}
\author[A]{\fnms{Sajad} \snm{Khodadadian}\ead[label=e1,mark]{skhodadadian3@gatech.edu}}
,\author[B]{\fnms{Thinh} \snm{T. Doan}\ead[label=e2,mark]{thinhdoan@vt.edu}} 
\author[B]{\fnms{Justin} \snm{Romberg}\ead[label=e3,mark]{jrom@ece.gatech.edu}}
\and
\author[A]{\fnms{Siva Theja} \snm{Maguluri}\ead[label=e4,mark]{siva.theja@gatech.edu}}
\address[A]{
Geogia Institute of Technology,
\printead{e1,e3,e4}}
\address[B]{
Virginia Tech,
\printead{e2}}
\end{aug}

\begin{abstract}
\input{abstract}
\end{abstract}
\end{frontmatter}

\input{intro}
\input{alg}
\input{main_result}
\input{need_for_ergodicity}
\input{proof_sketch}
\input{conclusion}

\bibliographystyle{imsart-number}
\bibliography{Refs}
\pagebreak
\begin{appendix}
\begin{center}{\huge \textbf{Appendices}}\end{center}
The supplementary material is organized as follows: in Section \ref{sec:C} the details of the Proof of Theorem \ref{thm: improv_mix} is presented, and in Section \ref{sec:Thm_main_pf} details of the proof of Proposition \ref{thm:main2} is provided.
\input{best_thm_proof}
\input{Thm_main_pf}
\end{appendix}
\end{document}

%% file: abstract.tex
Actor-critic style two-time-scale algorithms are one of the most popular methods in reinforcement learning, and have seen great empirical success.  However, their performance is not completely understood theoretically. In this paper, we characterize the \emph{global} convergence of an online natural actor-critic algorithm in the tabular setting using a single trajectory of samples. Our analysis applies to very general settings, as we only assume ergodicity of the underlying Markov decision process. In order to ensure enough exploration, we employ an $\epsilon$-greedy sampling of the trajectory.

For a fixed and small enough exploration parameter $\epsilon$, we show that the two-time-scale natural actor-critic algorithm has a rate of convergence of $\tilde{\mathcal{O}}(1/T^{1/4})$, where $T$ is the number of samples, and this leads to a sample complexity of $\Tilde{\mathcal{O}}(1/\delta^{8})$  samples to find a policy that is within an error of $\delta$ from the \emph{global optimum}. Moreover, by carefully decreasing the exploration parameter $\epsilon$ as the iterations proceed, we present an improved sample complexity of $\Tilde{\mathcal{O}}(1/\delta^{6})$ for convergence to the global optimum.

%% file: intro.tex
\section{Introduction}\label{sec:Intro}
In reinforcement learning (RL), an agent operating in an environment, modeled as a  Markov decision process (MDP), tries to learn a policy that maximizes its long-term reward. Methods for solving this optimization problem include value function methods, such as $Q$-learning \cite{watkins1992q}, and policy space methods, such as TRPO \cite{schulman2015trust}, PPO \cite{schulman2017proximal}, and actor-critic \cite{konda2000actor}.

Policy space methods explicitly search for the maximum of the value function $V^\pi$, which codifies the expected long-term reward, through iterative optimization over the policy $\pi$.  Although in general $V^\pi$ is a nonconvex function of $\pi$ \cite{agarwal2019theory}, global optimality can be obtained by employing either gradient descent \cite{agarwal2019theory,mei2020global}, mirror descent \cite{geist2019theory,shani2019adaptive}, or natural gradient descent \cite{agarwal2019theory}. These methods, however, assume access to an oracle that returns the gradient of the value function for any given policy.  In many practical scenarios, and in particular when the parameters of the MDP are only partially known, these gradients have to be estimated from observations or simulations.

Actor-critic (AC) techniques integrate estimation of the gradient into the policy search.  In this framework, a critic estimates the value ($Q$-function) of a policy, usually through a temporal difference iteration.  The actor then uses this estimate to form a gradient to improve the policy. AC algorithms have been observed to converge quickly relative to other methods \cite{wang2016sample, bahdanau2016actor}, and have enjoyed success in several applications including robotics \cite{Haarnoja_2019}, computer games \cite{espeholt2018impala}, and power networks \cite{Gajjar_2003}. 

AC algorithms can be classified into batch vs. online. In the batch setting, in each iteration of the AC, the critic evaluates the policy using a set of collected data. This type of batch update, however, cannot be implemented in an online manner, and requires simulations that need to be restarted in specific states, making its implementation appropriate in artificial environments such as Atari games \cite{schulman2017proximal}, but not in scenarios that require the agent to ``learn as they go''. 

A truly online and two-time-scale AC variant was first proposed in \cite{konda2000actor}, where at every iteration the actor and critic updates depend only on one sample observed from the environment using the current policy. Later \cite{peters2008natural,bhatnagar2009natural} presented a version of this algorithm using a natural policy gradient.  These methods can be viewed as two-time-scale stochastic approximation (SA) algorithms, where the actor and the critic operate at the ``slow'' and ``fast'' time scales, respectively. 

AC algorithms often use low-order approximations for the value and policy functions.  While this type of function approximation can dramatically simplify the learning process, thus allowing us to apply the algorithm to complex, real-world problems, these approximations introduce non-vanishing, systematic errors, as the truly optimal policy typically does not lie in the set of functions considered.  In this paper, however, we are completely focussed on recovering the globally optimal policy, and so we will operate in the ``tabular setting'' that considers every possible distribution over the (finite number of) actions for every possible state.

{\bf Main contributions:}
\begin{itemize}
    \item We analyze the two-time-scale natural AC algorithm in the tabular setting. Our setting can be seen as a two-time-scale linear stochastic approximation with a time-varying Markovian noise. Unlike several recent papers, we do not make an extensive set of assumptions. The \emph{only} assumption we make is the ergodicity of the underlying Markov chain under all policies.
    \item Our analysis show the importance of the exploration in AC type algorithms.
    We argue that a naive execution of natural AC fails to properly explore all of the state-action pairs, a fact that we illustrate with a simple example. Therefore, we employ $\epsilon$-greedy exploration to guarantee global convergence. 
    \item For a fixed and small enough exploration parameter $\epsilon$, we show that using $T$ number of samples, two-time-scale natural AC Algorithm \ref{alg:1} converges to within $\tilde{\mathcal{O}}\left(\frac{1}{\epsilon T^{1/4}}+\epsilon\right)$ of the global optimum. We show that for carefully chosen $\epsilon$, Algorithm \ref{alg:1} finds a policy within a $\delta$-ball around the global optimum using $\tilde{\mathcal{O}}(1/\delta^{8})$ samples.
    \item We show that using a time-varying $\epsilon$ improves the sample complexity of Algorithm \ref{alg:1} to $\tilde{\mathcal{O}}(1/\delta^{6})$.
\end{itemize}

\subsection{Related works}
\textbf{Stochastic approximation:} This method was first introduced in \cite{robbins1951stochastic}. Asymptotic convergence of stochastic approximation (SA) was studied in \cite{borkar2009stochastic, benveniste2012adaptive}. Recently, there has been a flurry of work on the finite time analysis of SA for both linear \cite{mou2020linear} and nonlinear \cite{chen2020finite, doan2021finite} operators, under both i.i.d \cite{liu2015finite} and Markovian \cite{kaledin2020finite} noise, with batch \cite{khodadadian2021finite} or two-time-scale \cite{borkar2009stochastic} updates. Our setting in this paper can be categorized as a linear two-time-scale SA with the noise generated from a time-varying Markov chain.

\textbf{Actor-critic:} AC algorithm was first proposed in \cite{konda2000actor} as a two-time-scale stochastic approximation \cite{borkar2008, borkar2018concentration, Kaledin_two_time_scale_2020, doan2019finite} variant of the policy gradient algorithm \cite{sutton2000policy}. In this algorithm a faster time scale is used to collect samples for gradient estimation, and a slower time scale is used to perform an update to the policy. In this paper we are interested in such a two-time-scale version of the natural policy gradient \cite{kakade2002natural}. While natural gradient descent is closely related to mirror descent  \cite{raskutti2015information,gunasekar2020mirrorless}, in the context of Markov decision processes they are known to be identical \cite{agarwal2019theory,geist2019theory}. Even though the objective $V^{\pi}$ is a nonconvex function of the policy $\pi$,  convergence rate of natural policy gradient to the global optimum under the planning setting (when the exact gradients are known) is recently established  in \cite{agarwal2019theory,geist2019theory,cen2020fast, cayci2021linear}. Natural AC, which is a variant of AC with natural policy gradient in the actor, was studied in \cite{NIPS2009_3767, NIPS2013_5184, peters2008natural,bhatnagar2009natural}. 

While the asymptotic convergence of  AC methods including natural AC is well-understood \cite{williams1990mathematical, borkar2009stochastic, konda2000actor, bhatnagar2009natural, zhang2020provably}, their finite-time convergence was largely unknown until recently \cite{shani2020adaptive, lan2021policy,zhang2021finite, khodadadian2021finite, zhang2019convergence, qiu2019finite, kumar2019sample, liu2019neural, wang2019neural,xu2020non, xu2020improving, liu2020improved, wu2020finite, chen2021finite, xu2021doubly}.
The authors in \cite{zhang2019convergence, qiu2019finite,kumar2019sample, xu2021doubly} provide the convergence rate of AC where the parameter of the critic is updated using a number of collected samples instead of only one single sample. Such a setting, referred to as batch AC, cannot be implemented in an online fashion since at any iteration the critic has to implement the current policy for a number of time steps to collect enough data. A similar batch approach was used in \cite{wang2019neural, xu2020improving, xu2020non, liu2020improved, lan2021policy,khodadadian2021finite, chen2021finite} to study natural AC and in \cite{liu2019neural, shani2020adaptive} to study the TRPO algorithm, which is a variant of mirror descent. The authors in \cite{khodadadian2021finite, chen2021finite} study the finite time convergence bound of off-policy natural AC algorithm under constant step size. However, due to the constant step size they do not have convergence to the global optimum. \cite{lan2021policy, zhan2021policy} study the finite time convergence of a regularized variant of natural AC with batch data update. In \cite{wu2020finite} the convergence of two-time-scale AC is analyzed. However, in \cite{wu2020finite} the convergence only to the local optimal is established.

In this paper, we study the original AC method \cite{konda2000actor} without considering  a batch update. In other words, data is collected through a single trajectory of a time-varying Markov chain and the update is performed in a two-time-scale manner. To the best of our knowledge, the only paper in the literature that considers such a setting is \cite{wu2020finite} which studies the AC algorithm under function approximation. Although their results are remarkable, they make several assumptions on the space of approximation functions. In Section \ref{subsec:noegreedy} we will explain why these assumptions cannot be satisfied in the tabular setting with zero approximation error. Another related work is \cite{xu2020improving} where the authors claim to have a single trajectory algorithm. However, as explained in \cite[Appendix C]{khodadadian2021finite}, the proposed algorithm in \cite{xu2020improving} is not single trajectory. 

\textbf{$\epsilon$-greedy: } One of the differences between our algorithm and the previous work is the inclusion of
$\epsilon$-greedy to the natural AC. This greedy step ensures sufficient exploration of our algorithm, while keeping the algorithm online. $\epsilon$-greedy \cite{montague1999reinforcement} is commonly employed in various settings such as $Q$-learning \cite{wunder2010classes}, multi-armed bandit \cite{kuleshov2014algorithms, tokic2010adaptive} and contextual bandit \cite{bouneffouf2012contextual}. In these algorithms, $\epsilon$-greedy is usually employed in order to ensure sufficient exploration \cite{montague1999reinforcement}. In this paper, we show that this greedy step can ensure the global convergence of the AC as well, and we characterize the rate of this convergence.

To summarize, the work in the literature over the last two decades has looked at various challenges thrown by AC algorithms under various assumptions and different simplifying models. This paper studies the greedy version of this algorithm and consequently in one place addresses several analytical challenges which include: (i) two-time-scale analysis, (ii) an online or single trajectory update, (iii) Markovian data samples, (iv) time-varying Markov chain, (v) asynchronous update in tabular setting, (vi) diminishing step sizes, and (vii) global convergence with minimal assumptions.

%% file: alg.tex
\section{Natural Actor-Critic Methods for Reinforcement Learning}\label{sec:NAC_method}
The environment of our RL problem is modeled by a Markov Decision Process (MDP) specified by $\mathcal{M} = (\mathcal{S},\mathcal{A},\mathcal{P},\mathcal{R},\gamma)$, where $\mathcal{S}$ and $\mathcal{A}$ are finite sets of states and actions, $\mathcal{P}$ is the set of  transition probability matrices,  $\gamma\in(0,1)$ is the discount factor, and $\mathcal{R}:\mathcal{S}\times\mathcal{A}\rightarrow[0,1]$ is the reward function, where without loss of generality we assume that the rewards are in $[0,1]$. We focus on randomized stationary policies, where each policy $\pi$ assigns to each state $s\in\mathcal{S}$ a probability distribution $\pi(\cdot\,|\,s)$ over $\mathcal{A}$. Each policy $\pi$ on the MDP, induces a Markov chain with transition probability $P^\pi(s'|s)=\sum_a \mathcal{P}(s'|s,a)\pi(a|s)$ on the states. Assuming that this Markov chain is irreducible, it induces a stationary distribution over states, which we denote by $\mu^\pi$. By definition, this distribution satisfies $(\mu^\pi)^T P^\pi=(\mu^\pi)^T$ \cite{hajek2015random}. 

For a fixed policy $\pi$, a sample trajectory of the states and actions is generated according to $S_{k+1}\sim \mathcal{P}(\cdot|S_k,A_k), A_{k+1}\sim \pi(\cdot|S_{k+1})$. The value function associated with $\pi$ and the state $s$ is defined as the expected discounted cumulative reward, i.e.$V^{\pi}(s)=\mathbb{E}\left[\sum_{k=0}^{\infty} \gamma^{k}\mathcal{R}(S_{k},A_{k}) \,|\,S_{0} = s, A_{k} \sim \pi(\cdot|S_{k})\right]$.
Furthermore, given an initial state distribution $\mathrm{P}$ over $\mathcal{S}$, we denote the expected cumulative reward for a policy $\pi$ as $V^{\pi}(\mathrm{P})$. 
The goal is to find a policy that maximizes this expected cumulative reward:
\begin{align}
    \label{sec:prob:obj_theta}
    \pi^*\in\argmax_{\pi} V^{\pi}(\mathrm{P}).
\end{align}
Throughout the paper, we denote $V^{\pi^*}$ as $V^*$.
It can be shown \cite{puterman1990markov} that the optimal policy $\pi^*$ is independent of the initial distribution $\mathrm{P}$, and hence throughout this paper we assume $\mathrm{P}$ as fixed and we denote $V^{\pi}\equiv V^{\pi}(\mathrm{P})$. It can be shown that value function can be written as $V^{\pi} = \sum_{s,a}d^{\pi}(s)\pi(a|s)\mathcal{R}(s,a)$, where $d^{\pi}$, denoted as the discounted state visitation \cite{sutton2000policy}, is defined as $d^{\pi}(s) = (1-\gamma)\sum_{k=0}^{\infty}\gamma^{k}P^{\pi}(S_{k} = s\,|\,s_{0}\sim\mathrm{P})$, 
with $P^{\pi}(S_{k} = s\,|\,S_{0}\sim\mathrm{P})$ being the probability that $S_k=s$ after executing policy $\pi$ starting from the initial distribution $\mathrm{P}$ at $k=0$.  Throughout, we denote $d^{\pi^*}_\mathrm{P}$ as $d^*$.

Given policy $\pi_t$, the Natural Policy Gradient (NPG) algorithm  \cite{kakade2002natural, agarwal2019theory} under the softmax parametrization updates the policy in every time step according to
\begin{equation}\label{eq:soft-max-update}
\pi_{t+1}(a|s)\hspace{-0.8mm}=\hspace{-0.8mm}\frac{\pi_{t}(a|s)\exp(\beta_tQ^{\pi_{t}}(s,a))}{\sum_{a'}\pi_{t}(a'|s)\exp(\beta_tQ^{\pi_{t}}(s,a'))}, \forall s,a,
\end{equation}
where  $Q^{\pi}(s,a) = \mathbb{E} [  \sum_{k=0}^{\infty} \gamma^{k}\mathcal{R}(S_{k},A_{k}) \,|\,S_{0} = s, A_{0} = a, A_k\sim\pi(\cdot|S_k)]$ is the $Q$-function corresponding to the policy $\pi$. Here $\beta_t$ is the step size which might be time dependent.

The update rule in \eqref{eq:soft-max-update} has multiple interpretations \cite{khodadadian2021linear}. Firstly, as explained in \cite{geist2019theory}, it can be seen as the update of the mirror descent for problem in \eqref{sec:prob:obj_theta} using negative entropy as the divergence generating function. Secondly, the NPG update in \eqref{eq:soft-max-update} can be seen as a pre-conditioned gradient ascent with softmax parameterization, where the pseudoinverse of the Fisher information matrix \cite{rissanen1996fisher} multiplies the gradient as a preconditioner \cite{kakade2002natural}. While mirror descent and natural gradient descent are distinct but related algorithms in general \cite{raskutti2015information,geist2019theory,gunasekar2020mirrorless}, they are both identical to \eqref{eq:soft-max-update} in our setting of solving the problem in \eqref{sec:prob:obj_theta} using softmax policy parametrization.

In the setting above, the NPG method finds a globally optimal policy with a provable rate; \cite{agarwal2019theory} shows that after $T$ iterations of the update \eqref{eq:soft-max-update} with constant step size $\beta_t=\beta$, it finds a policy whose expected cumulative reward is within $\mathcal{O}(1/T)$ of the optimal policy. The convergence bound in \cite{agarwal2019theory} is for the ``MDP setting'', where the $Q$-function is computed exactly for every candidate policy $\pi_t$. In the vast majority of reinforcement learning applications, however, $Q^{\pi_t}$ has to be estimated from simulations or observations.

\subsection{Two-Time-Scale Natural Actor-Critic Algorithm} \label{subsec:noegreedy}
In order to perform the NPG update \eqref{eq:soft-max-update} for an unknown environment, one can first estimate $Q^{\pi_t}$ using a batch of samples of state-action-rewards. However, using batch of data for the update of the $Q$-function has practical drawbacks. In particular, sampling of the batch data requires the state of the system to be reset frequently, which is not possible in environments such as robotics. A truly online, completely data-driven technique that keeps a running estimate of the $Q$-functions while performing NPG updates based on these estimates is presented in Algorithm \ref{alg:1} with $\epsilon_t=0$. In this algorithm, the ``critic'' implements an asynchronous update to the $Q$-function, where the only entry in the table that is changed at every iteration is the one corresponding to the observed state-action pair $(S_t,A_t)$. After this, the ``actor'' uses the estimated $Q$-table to update the policy using a natural policy gradient update of the form in \eqref{eq:soft-max-update}. The critic and the actor use different step sizes ($\alpha_t$ and $\beta_t$, respectively), a fact that is crucial to maintaining the algorithm's stability.

Due to the existence of two different step sizes, Algorithm \ref{alg:1} can be viewed as a variant of two-time-scale stochastic approximation \cite{borkar2008}. Intuitively, the critic has to collect information about the gradient at a faster time scale than the time scale at which the actor executes the gradient update --- in other types of policy gradient algorithms, this takes the form of multiple samples being generated in an inner loop.  Since the AC method performs both updates from a single sample, we can achieve a similar effect by having the actor take a more conservative step.

One of the main differentiators of our work with the existing literature on the convergence of AC algorithms is the update of the policy that mixes in a small multiple of the uniform distribution.  This mixing is necessary to ensure sufficient exploration of the state-action space. In this algorithm, at each iteration $t$, the action $A_{t+1}$ is sampled from the policy $\hat{\pi}_t$, which is a convex combination of the policy $\pi_t$ and the uniform distribution. This strategy ensures that the sampling policy $\hat{\pi}_t$ attains at least $\epsilon_t$ weight in all it's elements, even though some elements of $\pi_t$ might be arbitrary small. Furthermore, with introducing this step, we can ignore the technical assumptions which is made by the previous works. In Section \ref{sec:need_for_er_ex}, we give an example of an MDP with 4 states and 2 actions where a naive implementation of AC without this $\epsilon$-greedy exploration step results in a suboptimal policy. 

In the existing literature, this exploration is ensured through more stringent conditions on the problem structure, which \textit{if} satisfied, can guarantee enough \textit{exploration} by the AC algorithm (Assumption 1 in \cite{xu2020improving} and \cite{xu2020non}, Assumption 4.2 in \cite{liu2020improved}, Assumption 4.3 in \cite{fu2020single}, and Assumption 4.1 in \cite{wu2020finite}). These assumptions, however, need not necessarily be satisfied in the tabular MDP setting. In particular, all these assumptions require $\pi_t(a|s)$ to be bounded away from zero for all states and actions $s,a$ uniformly over time. However, we know that in an MDP, there always exist a deterministic policy which is a global optimal. This means that $\pi_t(a|s)$ can very likely go to zero for some sate-action pair $s,a$, and the assumption can be violated.

\begin{algorithm}[tb]\caption{Two-time-scale natural AC algorithm with $\epsilon$-greedy exploration}\label{alg:1}
\begin{algorithmic}
\STATE {\bfseries Input:} Iteration number $T>0$, step sizes $\alpha_t, \beta_t$, exploration parameter $\epsilon_t$, $Q_0(s,a)\in \mathbb{R}^{|\mathcal{S}||\mathcal{A}|}$, $\pi_0(a|s)=\hat{\pi}_0(a|s)=\frac{1}{|\mathcal{A}|},\forall s,a$.
\STATE Draw $S_0$ from some initial distribution and $A_0\sim \pi_0(\cdot|s_0)$\\ 
 \FOR{t=0,1,2,\dots,T}
  \STATE Sample $S_{t+1}\sim \mathcal{P}(\cdot|S_t,A_t), A_{t+1}
  \sim \pii_t(\cdot|S_{t+1})$
  \STATE $\alpha_t(s,a)=\alpha_t \mathbbm{1}\{S_t=s,A_t=a\},\forall s,a$
  \STATE $Q_{t+1}(s,a) = Q_t(s,a) + \alpha_t(s,a)  \big[\mathcal{R}(S_t,A_t) + \gamma Q_t(S_{t+1},A_{t+1})-Q_t(S_t,A_t)\big], \forall s,a$\label{eq:Qupdate}
  \STATE $\pi_{t+1}(a|s) = \pi_t(a|s)\frac{\exp(\beta_tQ_{t+1}(s,a))}{\sum_{a'}\pi_t(a'|s)\exp(\beta_tQ_{t+1}(s,a')) }, \forall s,a$
  \STATE $\pii_{t+1}=\frac{\epsilon_t}{|\mathcal{A}|}+(1-\epsilon_t)\pi_{t+1}$
 \ENDFOR
 \STATE Sample $\hat{T}$ from $\{0,1,\dots,T\}$ by distribution $P(\hat{T}=i)=\frac{\beta_i}{\sum_{j=0}^{T}\beta_j}$ 
 \STATE\textbf{Output:} $\hat{\pi}_{\hat{T}}$
\end{algorithmic}
\end{algorithm}

%% file: main_result.tex
\section{Main Result: Finite Time Convergence Bounds of Greedy Natural Actor-Critic}

In this section, we provide a finite-time performance guarantee for Algorithm \ref{alg:1}. In this algorithm, we can either choose a constant $\epsilon$-greedy parameter, or a time-varying one. The advantage of the constant $\epsilon$ is faster rate of convergence to a neighborhood of the optimal, and the advantage of the time-varying greedy parameter is global convergence without any necessary pre-specified error.

In order to characterize our convergence results, first we make the following assumption.
\begin{assumption}\label{ass:stat_pos}
For every deterministic policy $\pi$, the Markov chain induced by the transition probability $P^\pi$ is ergodic.
\end{assumption}
For more explanation regarding this assumption, look at Section \ref{sec:need_for_er_ex}.

We now present the main result of the paper. We bound the deviation of the value of the policy returned by \ref{alg:1} from the value of the optimal policy. 

\begin{theorem}\label{thm: improv_mix}
Suppose Assumption \ref{ass:stat_pos}  holds. Consider Algorithm \ref{alg:1} under the following step size parameters
\begin{align}
\alpha_t=\frac{\alpha}{(t+1)^\nu},  ~~   \beta_t=\frac{\beta}{(t+1)^\sigma},  ~~ \epsilon_t=\frac{\epsilon}{(t+1)^\xi},\label{eq:step_sizes}
\end{align}
with  $0\leq\xi<\nu<\sigma<1, ~\alpha,\epsilon\leq 1$. Then,  
\begin{align*}
  \E[V^*-V^{\pii_{\hat{T}}} ] \leq & \mathcal{O}(T^{\sigma-1})+\begin{cases}
\tilde{\mathcal{O}}(T^{-\sigma}) & \text{if}\quad 1> 2\sigma \\
\tilde{\mathcal{O}}(T^{\sigma-1}) & \text{o.w} 
\end{cases}+\begin{cases}
\tilde{\mathcal{O}}(\epsilon T^{\sigma-1}) & \text{if}\quad \xi+\sigma > 1\\
\tilde{\mathcal{O}}(\epsilon T^{-\xi}) & \text{o.w} 
\end{cases}\nonumber\\
&+ \begin{cases}\tilde{\mathcal{O}}(T^{0.5(\nu+\xi-1)}/\epsilon^{0.5})& \text{if}~~ \nu+\xi+1>2\sigma,\\
\tilde{\mathcal{O}}(T^{\sigma-1}/\epsilon^{0.5})& \text{o.w.}\end{cases}+\begin{cases}\tilde{\mathcal{O}}(t^{0.5(\xi-\nu)}/\epsilon^{0.5})& \text{if}~~ 2+\xi>\nu+2\sigma,\\
\tilde{\mathcal{O}}(T^{\sigma-1}/\epsilon^{0.5})& \text{o.w.}\end{cases}\nonumber\\&+ \begin{cases}\tilde{\mathcal{O}}(t^{-0.5})& \text{if}~~ 1>2\sigma,\\
\tilde{\mathcal{O}}(T^{\sigma-1})& \text{o.w.}\end{cases}+\begin{cases}\tilde{\mathcal{O}}(t^{0.5(\xi+\nu-2\sigma)}/\epsilon^{0.5})& \text{if}~~ 2+\xi+\nu>4\sigma,\\
\tilde{\mathcal{O}}(T^{\sigma-1}/\epsilon^{0.5})& \text{o.w,}\end{cases}\\
& + \begin{cases}\tilde{\mathcal{O}}(t^{\xi+\nu-\sigma}/\epsilon)& \text{if}~~ 2+2\nu+2\xi>4\sigma,\\
\tilde{\mathcal{O}}(T^{\sigma-1}/\epsilon)& \text{o.w.,}\end{cases}
\end{align*}
where $\tilde{\mathcal{O}}(\cdot)$ ignores the $\log (T)$ terms.
\end{theorem}
The proof of Theorem \ref{thm: improv_mix} is provided in Section \ref{sec:proofsketch}. Note that while $V^*$ is not random, but $V^{\hat{\pi}_T}$ is, since the policy $\hat{\pi}_T$ is a function of all the random variables drawn in Algorithm \ref{alg:1}.

Furthermore, we state two corollaries of Theorem \ref{thm: improv_mix} for different choices of $\xi$. 
\begin{corollary}\label{cor:1.1}
Suppose Assumption \ref{ass:stat_pos}  holds. Consider Algorithm \ref{alg:1} under the parameters in \eqref{eq:step_sizes}. Suppose $\xi=0$, $\nu=0.5$, and $\sigma=0.75$. We have:
\begin{align}\label{eq:cor1_b}
\E[V^*-V^{\pii_{\hat{T}}}]\leq \tilde{\mathcal{O}}\left(\frac{1}{\epsilon T^{1/4}}+\epsilon\right). 
\end{align}
Hence, the algorithm requires $\tilde{\mathcal{O}}(1/\delta^4)$ number of samples to get $\epsilon+\delta/\epsilon$ close to the global optimum. Furthermore, by taking $\epsilon=\mathcal{O}(\delta)$, we get $\E[V^*-V^{\pii_{\hat{T}}}]\leq \tilde{\mathcal{O}}(\delta)$ after $T=\tilde{\mathcal{O}}(1/\delta^{8})$ iteration of Algorithm \ref{alg:1}.
\end{corollary}
The sample complexity in Corollary \ref{cor:1.1} is relatively poor due to the $\frac{1}{\epsilon}$ term on the upper bound in \eqref{eq:cor1_b}, which is due to a constant exploration factor in AC. In the next corollary we show how to achieve a better sample complexity by gradually reducing the exploration factor $\epsilon_t$.
\begin{corollary}\label{cor:1.2}
Suppose Assumption \ref{ass:stat_pos}  holds. Consider Algorithm \ref{alg:1} under the parameters in \eqref{eq:step_sizes}. Suppose $\xi=1/6$, $\nu = 0.5$,  and $\sigma=5/6$. We have:
\begin{align*}
\E[V^*-V^{\pii_{\hat{T}}}]\leq \tilde{\mathcal{O}}(1/T^{1/6}). 
\end{align*}
In particular, we have $\E[V^*-V^{\pii_{\hat{T}}}]\leq \delta$ after $T=\tilde{\mathcal{O}}(1/\delta^{6})$ iterations of Algorithm $\ref{alg:1}$.
\end{corollary}
Corollaries \ref{cor:1.1} and \ref{cor:1.2} are direct application of Theorem \ref{thm: improv_mix}. In particular, in the case of $\xi=0$, the term $\epsilon T^{-\xi}$ in the bound of Theorem \ref{thm: improv_mix} will be a constant proportional to $\epsilon$, and the best rate of convergence is obtained by picking $\nu=0.5$ and $\sigma=0.75$ which gives Corollary \ref{cor:1.1}. Also assuming $\xi>0$, the best rate of convergence can be obtained by $\xi=1/6, \nu=0.5, \sigma=5/6$.

We should emphasize that Corollaries \ref{cor:1.1} and \ref{cor:1.2} characterize the sample complexity for global convergence of Algorithm \ref{alg:1} with the only assumption of ergodicity of the underlying MDP. This is indeed a much weaker assumption compared to the related work. 

%% file: need_for_ergodicity.tex
\section{Need for Exploration and Ergodicity}\label{sec:need_for_er_ex}
In this section we explain the necessity of the $\epsilon$-greedy step in Algorithm \ref{alg:1} and the ergodicity Assumption \ref{ass:stat_pos}. In iteration $t$ of the natural AC algorithm, the objective of the critic is to estimate the $Q$-function corresponding to the policy $\pi_t$. In two-time-scale natural AC, in each iteration $t$ the algorithm estimates the $Q$-function by updating only a single random element $(s=S_t, a=A_t)$ of the $Q_t$ table. In our analysis, the $\epsilon$-greedy step ensures that in each iteration of the algorithm, each of the actions are being sampled with probability at least $\epsilon_t$. Furthermore, Assumption \ref{ass:stat_pos} ensures that all the states of the MDP are visited infinitely often. In the following we show why both $\epsilon$-greedy and Assumption \ref{ass:stat_pos} are essential for the convergence of the natural AC algorithm.

\textbf{(I) $\epsilon$-greedy:} Following the update of the policy in Algorithm \ref{alg:1}, we have
\begin{align}
\pi_t(a|s)=&\frac{\exp(\sum_{l=0}^{t-1} \beta_{l}Q_{l+1}(s,a))}{\sum_{a'}\exp(\sum_{l=0}^{t-1} \beta_{l}Q_{l+1}(s,a'))}.\label{eq:pi_t_a_s2}
\end{align}
If for some state $s$, action $\Tilde{a}$ satisfies $Q_k(s,\tilde{a})\ll Q_k(s,a), ~\forall a\neq \tilde{a}$, by \eqref{eq:pi_t_a_s2}, $\pi_t(\tilde{a}|s)$ converges to zero geometrically. Thus, with high probability $(s,\tilde{a})$  will not be explored, and we might converge to a suboptimal policy. Note that the scenario explained here can very likely happen when $\mathcal{R}(s,\tilde{a})$ is negligible with respect to $\mathcal{R}(s,a)$ for other actions $a$. 

\begin{figure}
    \centering
    \includegraphics[width=80mm]{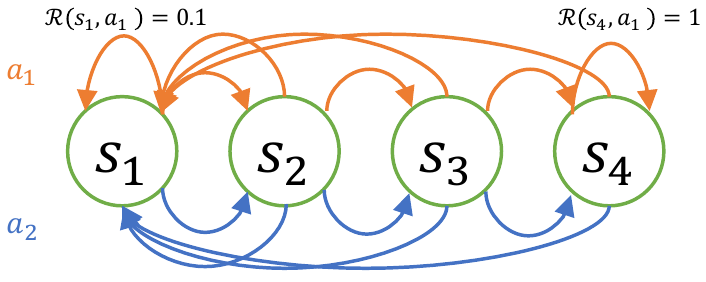}
    \caption{A 4 states, and 2 actions MDP. Orange and blue correspond to the non zero transition probabilities of actions $a_1$ and $a_2$ respectively.}
    \label{fig:non_explor_MDP}
\end{figure}

\begin{figure}
    \centering
    \includegraphics[width=80mm]{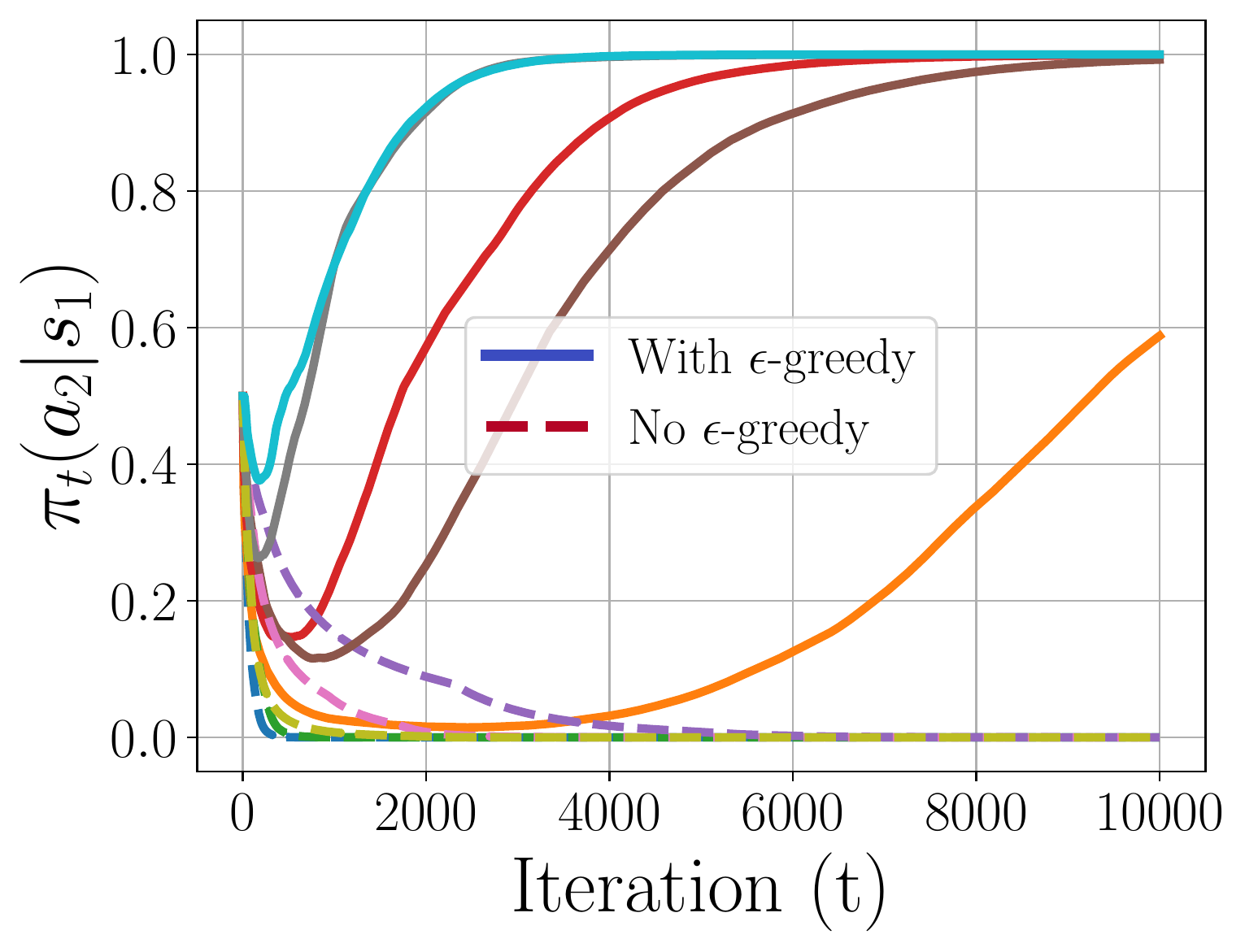}
    \caption{$\pi_t(a_2|s_1)$ for 10 trajectories generated by the natural AC algorithm over the MDP in Fig. \ref{fig:non_explor_MDP}. Straight and dashed lines show the result with and without $\epsilon$-greedy, respectively. Since here $\pi^*(a_2|s_1)=1$, it shows that the algorithm without $\epsilon$-greedy converges to a suboptimal policy.}
    \label{fig:pi_a1_s1222}
\end{figure}

The following experiment illustrates the necessity of the $\epsilon$-greedy policy update. Consider the MDP depicted in Fig. \ref{fig:non_explor_MDP}. This MDP has 4 states and 2 actions. All the transition probabilities depicted in the figure are positive, and the rest are zero. Furthermore, $\mathcal{R}(s_1,a_1)=0.1$ and $\mathcal{R}(s_4,a_1)=1$, and the rest of the rewards are zero. Suppose $\mathcal{P}(s_1|s_i,a_1)=0.999, ~i=1,2,3$, $\mathcal{P}(s_{i+1}|s_i,a_1)=0.001, ~i=1,2,3$, $\mathcal{P}(s_4|s_4,a_1)=0.999$, $\mathcal{P}(s_1|s_4,a_1)=0.001$, $\mathcal{P}(s_1|s_i,a_2)=0.001, ~i=2,3$, $\mathcal{P}(s_{i+1}|s_i,a_2)=0.999, ~i=2,3$, $\mathcal{P}(s_2|s_1,a_2)=1$, $\mathcal{P}(s_4|s_4,a_2) = 0.001$, $\mathcal{P}(s_1|s_4,a_2)=0.999$.
In this MDP, the optimal policy in state $s_1$ is to play action $a_2$. Fig. \ref{fig:pi_a1_s1222} shows $\pi_t(a_2|s_1)$ for 10 trajectories achieved by the natural AC. The straight lines show the output of the algorithm when $\epsilon$-greedy is employed, and the dashed lines are the output without $\epsilon$-greedy. It is clear that almost always the trajectories of the algorithm without $\epsilon$-greedy converge to a suboptimal policy. 

\textbf{(II) Ergodicity Assumption:} This assumption implies that under all policies, the induced Markov chain over the states of the MDP is irreducible and aperiodic. We discuss these two assumptions separately in the next two paragraphs. 

First, for an example of an MDP which does not satisfy irreducibility assumption, consider any episodic MDP, where there is a terminal state in which the episode ends \cite{montague1999reinforcement}. This system can be modeled as an infinite horizon MDP with an absorbing state corresponding to the terminal state. It is clear that this MDP does not satisfy irreducibility assumption. Furthermore, since after a finite time, with high probability we reach to the absorbing state, it is impossible to find the optimal policy using a \emph{single} trajectory.

Second, since here we assume finite state and action spaces, the aperiodicity assumption along with irreducibility is equivalent to the existence of a mixing time, which is common in the literature \cite{xu2020improving,xu2020non,fu2020single,wu2020finite,levin2017markov,lan2021policy}. We make this more precise in Lemma \ref{lem:mixing3}.

%% file: proof_sketch.tex
\section{Proof of Theorem \ref{thm: improv_mix}: Two-Time-Scale Analysis \label{sec:proofsketch}}

\input{proof_thm_1}

\subsection{Proof sketch of Lemma \ref{lem:b_t3}}\label{sec:lem_5.8_pf}
In Algorithm \ref{alg:1}, the actions $\{A_t\}_{t\geq1}$ are sampled from a time-varying policy $\pii_{t-1}$. Hence the tuple $(S_t,A_t)$ follows a time-varying Markov chain as follows
\begin{align*}
    S_{t-\tau}&\xrightarrow{\pii_{t-\tau-1}} A_{t-\tau}\xrightarrow{\mathcal{P}}S_{t-\tau+1}\xrightarrow{\pii_{t-\tau}} A_{t-\tau+1}\dots\xrightarrow{\mathcal{P}}S_t\xrightarrow{\pii_{t-1}}A_t\xrightarrow{\mathcal{P}}S_{t+1}\xrightarrow{\pii_{t}}A_{t+1}.
\end{align*}
Since the sampling policy is changing over time, the convergence analysis of this Markov chain is difficult. 

In order to analyze this time-varying Markov chain, at each time step $t$, we construct the following auxiliary Markov chain (This idea was first employed in \cite{Bhandari2018_FiniteTD}):
\begin{align*}
S_{t-\tau}&\xrightarrow{\pii_{t-\tau-1}} A_{t-\tau}\xrightarrow{\mathcal{P}}\Tilde{S}_{t-\tau+1}\xrightarrow{\pii_{t-\tau-1}} \Tilde{A}_{t-\tau+1}\dots\xrightarrow{\mathcal{P}}\Tilde{S}_t\xrightarrow{\pii_{t-\tau-1}}\Tilde{A}_t\xrightarrow{\mathcal{P}}\Tilde{S}_{t+1}\xrightarrow{\pii_{t-\tau-1}}\Tilde{A}_{t+1}. 
\end{align*}
Due to the geometric mixing of the Markov chain, which is stated formally in Lemma \ref{lem:mixing3}, by choosing $\tau$ large enough, the distribution of $(\tilde{S}_{t+1},\tilde{A}_{t+1})$ is ``sufficiently close'' to the stationary distribution $\mu^{\pii_{t-\tau-1}}\otimes\pii_{t-\tau-1}$. 
\begin{lemma}\label{lem:mixing3}
Suppose Assumption \ref{ass:stat_pos} holds for an MDP. Then there exist $m>0$ and $\rho\in(0,1)$, such that
\begin{align}
d_{TV}(\mu^\pi(\cdot),P^\pi(S_\tau=\cdot|S_1=s))\leq m\rho^{\tau}, \forall s\in\mathcal{S},\forall \pi,\label{eq:mixing3}
\end{align}
where $d_{TV}(\cdot,\cdot)$ denotes the total variation distance between two distributions. Furthermore, aperiodicity and the existence of $m$ and $\rho$ in inequality \eqref{eq:mixing3} are equivalent, i.e., if there exist a policy such that the underlying Markov chain is periodic, then \eqref{eq:mixing3} does not hold. 
\end{lemma}

Define $\tilde{O}_t=(\Tilde{S}_t,\Tilde{A}_t,\Tilde{S}_{t+1},\Tilde{A}_{t+1})$. We have:
\begin{align}
    \Gamma(\pii_{t-1},\theta_t,O_t)=&\Gamma(\pii_{t-1},\theta_t,O_t) - \Gamma(\pii_{t-\tau-1},\theta_t,O_t)\label{eq:pi_lip}\\
    &+\Gamma(\pii_{t-\tau-1},\theta_t,O_t)-\Gamma(\pii_{t-\tau-1},\theta_{t-\tau},O_t)\label{eq:theta_lip}\\
    &+\Gamma(\pii_{t-\tau-1},\theta_{t-\tau},O_t)-\Gamma(\pii_{t-\tau-1},\theta_{t-\tau},\tilde{O}_t)\label{eq:OvsOtilde}\\
    &+\Gamma(\pii_{t-\tau-1},\theta_{t-\tau},\tilde{O}_t)\label{eq:Otilde}.
\end{align} 
We bound each of the terms above separately. 
Due to the Lipschitzness of $\Gamma$ with respect to it's first and second arguments, the terms \eqref{eq:pi_lip} and \eqref{eq:theta_lip} can be bounded  as follows:
\begin{align*}
\Gamma(\pii_{t-1},\theta_t,O_t) -\Gamma(\pii_{t-\tau-1},\theta_t,O_t)  \leq& \mathcal{O}\left(\|\pii_{t-1}-\pii_{t-\tau-1}\|\right)\\
\leq& \mathcal{O}\left(\sum_{i=t-\tau}^{t-1}\left\|\pii_i-\pii_{i-1}\right\|\right)
\leq \mathcal{O}\left(\tau\frac{\epsilon_{t}}{t}+\tau\beta_{t}\right).
\end{align*}
\begin{align*}
\Gamma(\pii_{t-\tau-1},\theta_t,O_t) -\Gamma(\pii_{t-\tau-1},\theta_{t-\tau},O_t)\leq&  \mathcal{O}\left(\|\theta_t-\theta_{t-\tau}\|\right)\\
\leq&  \mathcal{O}\left(\sum_{i=t-\tau+1}^t\|\theta_i-\theta_{i-1}\|\right)
\leq  \mathcal{O} \left(\tau\alpha_{t}+\tau\frac{\epsilon_{t}}{t} +  \tau\beta_{t}\right).
\end{align*}
In order to bound the remaining two terms \eqref{eq:OvsOtilde} and \eqref{eq:Otilde}, we first apply conditional expectation on both sides. Bounding \eqref{eq:OvsOtilde} is slightly technical and is presented in Lemma \ref{lem:D.6}. The main idea is as follows. Since the policy $\pii_{t}$ does not change very fast over time, the conditional expectation of $\Gamma(\pii_{t-\tau-1},\theta_{t-\tau},O_t)$ and $\Gamma(\pii_{t-\tau-1},\theta_{t-\tau},\tilde{O}_t)$ are close. Denoting $\bar{\mathcal{F}}_{t-\tau}=\{S_{t-\tau},\pii_{t-\tau-1},\theta_{t-\tau}\}$, we have
\begin{align*}
&\E[\Gamma(\pii_{t-\tau-1},\theta_{t-\tau},O_t) -\Gamma(\pii_{t-\tau-1},\theta_{t-\tau},\tilde{O}_t)|\bar{\mathcal{F}}_{t-\tau}] \leq \tilde{\mathcal{O}} \left(\sum_{i=t-\tau}^t\|\pii_i-\pii_{t-\tau-1}\|\right) \leq \mathcal{O}\left(\tau\frac{\epsilon_{t}}{t}+\tau\beta_{t}\right).
\end{align*}
Finally, denoting $O'_t=(S_t',A_t',S_{t+1}',A_{t+1}')$, where $S_t'\sim \mu^{\pii_{t-\tau-1}}$, $A_t'\sim \pii_{t-\tau-1}(\cdot|S_t')$, $S_{t+1}'\sim \mathcal{P}(\cdot|S_t',A_t')$, and $A_{t+1}'\sim \pii_{t-\tau-1}(\cdot|S_{t+1}')$, we have $\E  \left[\Gamma(\pii_{t-\tau-1},\theta_{t-\tau},O'_t)\big|S_{t-\tau},\pii_{t-\tau-1}\right] = 0$ due to the Bellman equation. According to Lemma \ref{lem:mixing3}, the distribution of the auxiliary chain $\tilde{O}_t=(\Tilde{S}_t,\Tilde{A}_t,\Tilde{S}_{t+1},\Tilde{A}_{t+1})$ converges geometrically fast to the distribution of $O'_t=(S_t',A_t',S_{t+1}',A_{t+1}')$. Hence, we have
\begin{align*}
\E\left[\Gamma(\pii_{t-\tau-1},\theta_{t-\tau},\Tilde{O}_t)\big|S_{t-\tau},\pii_{t-\tau-1}\right] &\leq \mathcal{O}(\rho^{\tau}).
\end{align*}
Putting all the above bounds together, we get the result. \QEDopen

\subsection{Explanation of the main lemmas} \label{sec:lems_explan}

Lemma \ref{lem:cont3} which provides a bound on one step drift of the $Q$-function with respect to the sampling policy $\pii_t$ can be derived from Lemmas \ref{lem:cont4} and \ref{lem:pi_lip3} below.
\begin{lemma}\label{lem:cont4}
For every pair of policies $\pi_1$ and $\pi_2$, we have:
\[
\|Q^{\pi_1}-Q^{\pi_2}\| \leq \CLL\|\pi_1-\pi_2\|,
\]
where $\CLL =\frac{\gamma|\mathcal{S}||\mathcal{A}|}{(1-\gamma)^2}$. 
\end{lemma}

\begin{lemma}\label{lem:pi_lip3}
The policy $\pii_t$, satisfies the following:
\begin{align*}
\|\pii_{t+1}-\pii_t\|&\leq \CL \beta_t + \CLLL \frac{\epsilon_{t-1}}{t}, ~\forall t\geq 1,
\end{align*}
where $\CL=Q_{\max}\sqrt{|\mathcal{A}||\mathcal{S}|}$ and $\CLLL=\xi\sqrt{|\mathcal{S}|}(\frac{1}{\sqrt{|\mathcal{A}|}}+1)$.
\end{lemma}
Lemma \ref{lem:cont4} characterizes the Lipschitzness of the $Q^\pi$ function with respect to the policy $\pi$, and Lemma \ref{lem:pi_lip3} provides an upper bound on the drift of the sampling policy $\pii_t$.

Finally, Lemma \ref{lem:mulem2} below provides an intuition regarding the constant $\mu$ in Lemma \ref{lem:mu3}.
\begin{lemma}\label{lem:mulem2}
Suppose Assumption \ref{ass:stat_pos} holds. There exist a constant $\mu>0$ such that for all the policies $\pi$, the stationary distribution $\mu^\pi$ satisfies 
\[
\mu^\pi(s)\geq \mu, \forall s\in\mathcal{S}.
\]
\end{lemma}
Lemma \ref{lem:mulem2} is a direct consequence of the ergodicity of the underlying MDP. In particular, the ergodicity Assumption \ref{ass:stat_pos} ensures that for all the policies $\pi$, under the stationary distribution $\mu^\pi$, all the states are being visited with rate at least $\mu$. As explained in Section \ref{sec:need_for_er_ex} this is indeed essential for the convergence of AC algorithm.

%% file: proof_thm_1.tex
Next we provide the proof of Theorem \ref{thm: improv_mix}. Before expressing the proof, we state the following Proposition on the convergence of the natural AC along with its proof.
\begin{proposition}\label{thm:main2}
Consider the two-time-scale natural AC Algorithm \ref{alg:1} with $T$ iterations, and the output $\pii_{\hat{T}}$. Suppose the step size $\beta_t$ and the exploration parameter $\epsilon_t$ are non-increasing with respect to $t$. We have the following:
\begin{align*}
    \E[V^*-V^{\pii_{\hat{T}}} ] \leq\frac{1}{\sum_{t=0}^T\beta_t}\left\{\frac{2\beta}{(1-\gamma)^2}\hspace{-0.5mm}+\hspace{-0.5mm}\frac{\log|\mathcal{A}|}{(1-\gamma)}\hspace{-0.5mm}+\hspace{-0.5mm}\frac{2}{1-\gamma}\sum_{t=0}^T \bigg[  \beta_t\E\|Q^{\hat{\pi}_t}-Q_{t+1}\|+\frac{\CL\sqrt{|\mathcal{A}|}}{(1-\gamma)}\beta_t^2 + \epsilon_{t}\beta_t\bigg]\right\},
\end{align*}
where $\|\cdot\|$ is the euclidean norm and $\CL$ is a constant whose precise value is given in Lemma \ref{lem:pi_lip3} in Section \ref{sec:lems_explan}. 
\end{proposition}

\subsection{Proof of Proposition \ref{thm:main2}}
In this section we provide the proof of Proposition \ref{thm:main2}. A similar result was proved for NPG in \cite{agarwal2019theory}, when the actor has access to the exact $Q$-function. However, since the actor in Algorithm \ref{alg:1} has only access to $Q_{t}(s,a)$, rather than the exact $Q$-function $Q^{\pi_t}(s,a)$, establishing the bound in Proposition \ref{thm:main2} is more challenging. Note that  $Q_{t}(s,a)$ is obtained by the critic carrying out only one step of the TD-learning 
using a single sample update at each time step. Consequently, the error bound in Proposition \ref{thm:main2} involves the term $\frac{1}{T}\sum_{t=0}^T   \beta_t\E\|Q^{\hat{\pi}_t}-Q_{t+1}\|$, which  accounts for the time-average error in the critic's estimate of the $Q$-function. Proposition \ref{thm:main2} is also different from the results in \cite{agarwal2019theory} is terms of the step size. In particular, while \cite{agarwal2019theory} only considers the case of constant step size, the result in Proposition \ref{thm:main2} is stated for general choice of non-increasing step sizes. Furthermore, a similar type of upper bound has been established in \cite{khodadadian2021finite, chen2021finite} for the analysis of off-policy natural AC. However, in those works the $\epsilon_T$ term is absent.

When the actor has access to the exact $Q$-functions, it was observed in \cite{agarwal2019theory} that using a constant step size results in $\mathcal{O}(1/T)$ rate of convergence. This result can be reproduced from Proposition \ref{thm:main2} by eliminating the $\frac{1}{T}\sum_{t=0}^T   \beta_t\E\|Q^{\hat{\pi}_t}-Q_{t+1}\|$ term in the upper bound, and taking a constant $\beta_t$, and choosing $\epsilon_T=0$. However, due to the existence of $\epsilon_T$, and the $\frac{1}{T}\sum_{t=0}^T   \beta_t\E\|Q^{\hat{\pi}_t}-Q_{t+1}\|$ term, the optimal convergence rate can only be obtained by a carefully diminishing step size, which has been shown in Theorem \ref{thm: improv_mix}.

Next we provide the proof of Proposition \ref{thm:main2}. 

\textit{Proof of Proposition \ref{thm:main2}:}
We will use a Lyapunov drift based argument to prove the proposition using the KL-divergence \cite{cover2012elements} as a Lyapunov or potential function. This is a natural choice because it is known to be the right potential function for mirror descent \cite{bansal2019potential} in optimization and it is known \cite{agarwal2019theory,geist2019theory} that the natural gradient ascent is equivalent to mirror descent. 

Let $M(\pi)=\E_{s\sim d^*}[D_{KL}(\pi^*(\cdot|s)||\pi(\cdot|s))]$. Then, \begin{align*}
    M(\pi_{t+1})-M(\pi_t)
    =&\E_{s\sim d^*}\left[ \sum_a\pi^*(a|s)\log\frac{\pi_t(a|s)}{\pi_{t+1}(a|s)}\right]\\
    \stackrel{(a)}{=}&\sum_{s,a}d^*(s)\pi^*(a|s)(\log Z_t(s)-\beta_t Q_{t+1}(s,a))\\
    \stackrel{(b)}{=}&\beta_t\sum_{s,a}d^*(s)\pi^*(a|s)(Q^{\hat{\pi}_t}(s,a)-V^{\hat{\pi}_t}(s))\\ 
    &+\beta_t\sum_{s,a}d^*(s)\pi^*(a|s)\left(Q_{t+1}(s,a)-Q^{\hat{\pi}_t}(s,a)- Q_{t+1}(s,a) + V^{\hat{\pi}_t}(s)\right)\\
    &+\sum_{s,a}d^*(s)\pi^*(a|s)(\log Z_t(s)-\beta_tQ_{t+1}(s,a))\\
    \stackrel{(c)}{=}&(1-\gamma)\beta_t\left[V^{\hat{\pi}_t}-V^{*}\right]+\beta_t\sum_{s,a}d^*(s)\pi^*(a|s)\left[Q^{\hat{\pi}_t}(s,a)-Q_{t+1}\right]\\
    &+\sum_{s,a}d^*(s)\left[\log Z_t(s)-\beta_tV^{\hat{\pi}_t}\right],
\end{align*}
where, $(a)$ is by the update rule of the policy $\pi_t$ in Algorithm \ref{alg:1} with $Z_{t}(s) = \sum_{a}\pi_t(a|s)\exp(\beta_tQ_{t+1}(s,a))$, $(b)$ is by adding and subtracting terms, and $(c)$ is by Performance Difference Lemma \cite{Kakade02approximatelyoptimal}. Rearranging the terms, we get:
\begin{align}
V^{*}\vspace{-1mm}-V^{\hat{\pi}_t}
=&\frac{1}{(1-\gamma)\beta_t}\left[M(\pi_t)-M(\pi_{t+1})\right]\nonumber\\
&+\frac{1}{1-\gamma}\sum_{s,a}d^*(s) \pi^*(a|s)\left[Q^{\hat{\pi}_t}(s,a)-Q_{t+1}(s,a)\right]\label{eq:Q_diff3}\\
&+\frac{1}{1-\gamma}\sum_{s,a}d^*(s)\left[\frac{1}{\beta_t}\log Z_t(s)-V^{\hat{\pi}_t}(s)\right].\label{eq:betaZ_t3}
\end{align}
We bound the terms in \eqref{eq:Q_diff3} and \eqref{eq:betaZ_t3} separately. 

Using the Cauchy-Schwarz inequality in \eqref{eq:Q_diff3}, we have: 
\begin{align}
    \frac{1}{1-\gamma}\sum_{s,a}d^*(s) \pi^*(a|s)\left[Q^{\hat{\pi}_t}(s,a)-Q_{t+1}(s,a)\right]
    \leq &\frac{1}{1-\gamma}\|Q^{\hat{\pi}_t}-Q_{t+1}\|.\nonumber
\end{align}
In order to bound the term \eqref{eq:betaZ_t3}, we use the following lemma.
\begin{lemma}\label{lem:logZ_bound3}
Consider Algorithm \ref{alg:1} with  $Z_{t}(s) = \sum_{a}\pi_t(a|s)\exp(\beta_tQ_{t+1}(s,a))$. For any $t\geq 0$ we have the following inequality:
\begin{align*}
    \sum_{s}d^*(s)\left[\frac{1}{\beta_t}\log Z_t(s)-V^{\hat{\pi}_t}(s)\right]\nonumber 
    \leq V^{\pi_{t+1}}(d^*)-V^{\pi_{t}}(d^*) + \|Q^{\pii_t}-Q_{t+1}\| + \frac{2\CL\sqrt{|\mathcal{A}|}}{(1-\gamma)^2}\beta_t + \frac{ \epsilon_{t-1}}{1-\gamma},
\end{align*}
where $\epsilon_{-1}:=0$.
\end{lemma}
Employing Lemma \ref{lem:logZ_bound3} in \eqref{eq:betaZ_t3}, multiplying by $\beta_t$ and summing from $0$ to $T$, we get:
\begin{align}
\sum_{t=0}^T \beta_t(V^{*}-V^{\pii_t})
\leq & \sum_{t=0}^T \frac{1}{(1-\gamma)}\left[M(\pi_t)-M(\pi_{t+1})\right]\nonumber\\
&+ \frac{2\beta_t}{1-\gamma}\|Q^{\hat{\pi}_t}-Q_{t+1}\|+\frac{\beta_t}{1-\gamma}\left[V^{\pi_{t+1}}(d^*)-V^{\pi_{t}}(d^*)\right] \nonumber\\
&+ \frac{2\CL\sqrt{|\mathcal{A}|}}{(1-\gamma)^2}\beta_t^2 + \frac{\epsilon_{t-1}\beta_t}{1-\gamma} \nonumber\\
=&\frac{1}{(1-\gamma)}\sum_{t=0}^T \bigg\{M(\pi_t)-M(\pi_{t+1}) +\beta_t\left[ V^{\pi_{t+1}}(d^*)-V^{\pi_{t}}(d^*)\right]\bigg\}\label{eq:I14p}\\
&+\sum_{t=0}^T \bigg\{  \frac{2\beta_t}{1-\gamma}\|Q^{\hat{\pi}_t}-Q_{t+1}\|+\frac{2\CL\sqrt{|\mathcal{A}|}}{(1-\gamma)^2}\beta_t^2+ \frac{\epsilon_{t-1}\beta_t}{1-\gamma}\bigg\}.\label{eq:I1I4p}
\end{align}
We evaluate \eqref{eq:I14p} and \eqref{eq:I1I4p} separately. First, we have:
\begin{align}
\eqref{eq:I14p} =&\frac{1}{(1-\gamma)}\sum_{t=0}^T \bigg[\beta_t\left[ V^{\pi_{t+1}}(d^*)-V^{\pi_{t}}(d^*)\right]\bigg]+ \frac{1}{(1-\gamma)}\left[M(\pi_0)-M(\pi_{T+1})\right]\nonumber\\
\stackrel{(a)}{\leq}&\frac{1}{(1-\gamma)}\sum_{t=0}^T \bigg[\beta_t\left[ V^{\pi_{t+1}}(d^*)-V^{\pi_{t}}(d^*)\right]\bigg] + \frac{\log|\mathcal{A}|}{(1-\gamma)}\nonumber\\
=&\frac{1}{(1-\gamma)}\sum_{t=0}^T\left[(\beta_{t}-\beta_{t+1})V^{\pi_{t+1}}(d^*)\right]-\frac{\beta_0}{(1-\gamma)}V^{\pi_{0}}(d^*)+\frac{\beta_{T+1}}{(1-\gamma)}V^{\pi_{T+1}}(d^*)+ \frac{\log|\mathcal{A}|}{(1-\gamma)}\nonumber\\
\leq &\frac{1}{(1-\gamma)}\sum_{t=0}^T\left[(\beta_{t}-\beta_{t+1})V^{\pi_{t+1}}(d^*)\right]+\frac{\beta_{T+1}}{(1-\gamma)}V^{\pi_{T+1}}(d^*) + \frac{\log|\mathcal{A}|}{(1-\gamma)}\nonumber\\
\stackrel{(b)}{\leq}& \frac{1}{(1-\gamma)^2}\sum_{t=0}^T(\beta_{t}-\beta_{t+1})+\frac{\beta_{T+1}}{(1-\gamma)^2} + \frac{\log|\mathcal{A}|}{(1-\gamma)}\nonumber\\
= &\frac{1}{(1-\gamma)^2}(\beta_0-\beta_{T+1}) + \frac{\beta_{T+1}}{(1-\gamma)^2} + \frac{\log|\mathcal{A}|}{(1-\gamma)} \nonumber\\
\leq &\frac{2\beta}{(1-\gamma)^2}+\frac{\log|\mathcal{A}|}{(1-\gamma)},\label{eq:I_116p}
\end{align}
where $(a)$ is due to $0\leq D_{KL}(P(X)||Unif(X))\leq\log |\mathcal{X}|$, where $P(X)$ is any distributieseseon over $\mathcal{X}$ and $Unif(X)$ is uniform distribution over $\mathcal{X}$, and $|\mathcal{X}|$ is the cardinality of the random variable $X$\cite{cover2012elements}, and $(b)$ is due to Lemma \ref{lem:bounds3} and $\beta_t$ being non-increasing with respect to $t$.

Furthermore, we have
\begin{align}
\eqref{eq:I1I4p}
=&\frac{2}{1-\gamma}\sum_{t=0}^T \Bigg[  \beta_t\|Q^{\hat{\pi}_t}-Q_{t+1}\|+\frac{\CL\sqrt{|\mathcal{A}|}}{(1-\gamma)}\beta_t^2 + 0.5\epsilon_{t-1}\beta_t\Bigg]\nonumber\\
\leq&\frac{2}{1-\gamma}\sum_{t=0}^T \Bigg[  \beta_t\|Q^{\hat{\pi}_t}-Q_{t+1}\|+\frac{\CL\sqrt{|\mathcal{A}|}}{(1-\gamma)}\beta_t^2 + \epsilon_{t}\beta_t\Bigg].\label{eq:39and40}
\end{align}
Dividing both sides of \eqref{eq:I_116p} and \eqref{eq:39and40} with $\sum_{t=0}^T\beta_t$, and noting that $\frac{\sum_{t=0}^T\beta_t\E(V^*-V^{\hat{\pi}_t})}{\sum_{t=0}^T}=\E [V^*-V^{\hat{\pi}_{\hat{T}}}]$, we get the proposition.\QED

According to Proposition \ref{thm:main2}, to establish a bound for the performance metric $\E[V^*-V^{\pii_{\hat{T}}}]$, we need to characterize a bound for $\E\|Q^{\hat{\pi}_t}-Q_{t+1}\|$ for all $0\leq t\leq T$. Next we provide the proof of Theorem \ref{thm: improv_mix} which essentially corresponds to the characterization of this bound.
\subsection{Proof of Theorem \ref{thm: improv_mix}}
First, we introduce some notations and lemmas which will be used within the proof.
\begin{align}
    O_t &= (S_t,A_t,S_{t+1},A_{t+1})\nonumber\\
    r(O_t) &= [0;0;\dots;0;\mathcal{R}(S_t,A_t);0;\dots;0]\in \mathbb{R}^{|\mathcal{S}||\mathcal{A}|}\nonumber\\
    A(O)&\in\mathbb{R}^{|\mathcal{S}||\mathcal{A}|\times |\mathcal{S}||\mathcal{A}|}\nonumber\\ A(O)_{i,j}&\equiv A(s,a,s',a')_{i,j}
    =\begin{cases}\gamma-1& i=j=(s,a)=(s',a')\\-1&  (s,a)\neq(s',a'), i=j=(s,a)\\\gamma& (s,a)\neq(s',a'),  i=(s,a), j=(s',a')\\0 & \text{otherwise}
    \end{cases}\nonumber\\
    \theta_t&=Q_{t} - Q^{\pii_{t-1}}\label{eq:theta_t_def}\\
    \Bar{A}^\pi=&\E_{s\sim\mu^\pi(\cdot),a\sim\pi(\cdot|s),s'\sim \mathcal{P}(\cdot|s,a),a'\sim\pi(\cdot|s')}[A(s,a,s',a')]\label{eq:A_bar_pi_def}\\
    \Gamma(\pi,\theta,O)&=\theta^\top (r(O)+A(O)Q^\pi)+\theta^\top (A(O)-\Bar{A}^\pi)\theta.\nonumber
\end{align}
Note that with the above notation, the update of the $Q$-function in Algorithm \ref{alg:1} in the vector form can be written as: 
\[
Q_{t+1}=Q_t+\alpha_t(r(O_t)+A(O_t)Q_t),
\]
which by adding and subtracting terms, can be equivalently written as:
\[
\theta_{t+1}+(Q^{\pii_{t}}-Q^{\pii_{t-1}})=\theta_t+\alpha_t(r(O_t)+A(O_t)Q^{\pii_{t-1}}+A(O_t)\theta_t).
\]
Lemmas \ref{lem:cont3} and \ref{lem:D_dif_square3} characterize an upper bound on the one step drift of $Q^{\pii_t}$ and $\theta_t$.
\begin{lemma}\label{lem:cont3}
One step drift of the $Q$-function with respect to the sampling policy $\pii_t$ satisfies the following:
\[
\|Q^{\pii_{t+1}}-Q^{\pii_t}\| \leq \CLL (\CLLL \frac{\epsilon_{t-1}}{t} + \CL \beta_t),
\]
where the constants $\CL$, $\CLL$, and $\CLLL$ are defined in Lemmas \ref{lem:cont4} and \ref{lem:pi_lip3}. 
\end{lemma}
\begin{lemma}\label{lem:D_dif_square3}
The one step drift of the vector $\theta_t$ can be bounded as
\[
\|\theta_{t+1}-\theta_t\|^2\leq 2\alpha_t^2\Delta_Q^2+4\CLL^2\CLLL^2\frac{\epsilon_{t-2}^2}{(t-1)^2}+4\CLL^2\CL^2\beta_{t-1}^2,
\]
where $\Delta_Q$ is defined in Lemma \ref{lem:bounds} in the Appendix.
\end{lemma}
The following lemma is directly used to create a negative drift, which is essential for the convergence proof of Theorem \ref{thm: improv_mix}.

\begin{lemma}\label{lem:mu3}
Consider the policy $\pii_{t-1}$ in the $t-1$'th iteration of Algorithm \ref{alg:1}, and the vector $\theta_t$ and the matrix $\bar{A}^{\pii_{t-1}}$ defined in \eqref{eq:theta_t_def} and \eqref{eq:A_bar_pi_def}, respectively. We have:
\[
\theta^\top_t \bar{A}^{\pii_{t-1}} \theta_t \leq -(1-\gamma)\frac{\epsilon_{t-2}}{|\mathcal{A}|}\mu\|\theta_t\|^2,
\]
where $\mu>0$ is some absolute constant. Later in Lemma \ref{lem:mulem2} we explain the intuition behind the constant $\mu$.
\end{lemma}
The following Lemma provides some absolute bounds on the value and $Q$-function.
\begin{lemma}\label{lem:bounds3}
Let $Q_{\max}=\frac{1}{1-\gamma}$. Then we have
\begin{enumerate}
    \item  $0\leq V^\pi\leq Q_{\max}$
    \item $0\leq Q^\pi(s,a)\leq Q_{\max}$
    \item $\|Q^\pi\|\leq \sqrt{|\mathcal{S}||\mathcal{A}|}Q_{\max}$
    \item $\|Q_t\|\leq \sqrt{|\mathcal{S}||\mathcal{A}|}Q_{\max}$.
\end{enumerate}
\end{lemma}
A major part of the proof of Theorem \ref{thm: improv_mix} is to establish a bound on $\E[\Gamma(\pii_{k-1},\theta_k,O_k)]$. 
In the following, we provide such a bound in Lemma \ref{lem:b_t3}. The proof of this lemma is provided in Section \ref{sec:lem_5.8_pf}.

\begin{lemma} \label{lem:b_t3}
For any $\tau< t$, we have:
\begin{align}
\E\big[\Gamma(\pii_{t-1},\theta_t,O_t)\big] \leq& C_bm\rho^{\tau} +K_2\Delta_Q\tau\alpha_{t-\tau} +(C_u\CLLL+K_1\CLLL+K_2\CLL \CLLL)\frac{(\tau+1)^2\epsilon_{t-\tau-2}}{t-\tau-1}\nonumber\\
& + (C_u\CL+K_1\CL+K_2\CL \CLL) (\tau+1)^2 \beta_{t-\tau-1}.\nonumber
\end{align}
\end{lemma}

We further define
\begin{align}\label{eq:tau_t}
\tau_t:=\min \left\{r > 0|M_\rho\rho^{r}\leq\beta_t, r ~ \text{integral}\right\},\end{align}
where $M_\rho=(-\sigma/\ln(\rho))^\sigma/(\rho^{1+\sigma/\ln(\rho)})$.  It is easy to see that $1\leq\tau_t\leq t$ for all $t$, and $\tau_t=\mathcal{O}(\log(t))=\tilde{\mathcal{O}}(1)$.

In order to establish the convergence result in Theorem \ref{thm: improv_mix}, we use the bound in Proposition \ref{thm:main2}. By the definition of the step sizes, it is clear that $\sum_{t=0}^T\beta_t=\Theta(T^{1-\sigma})$. Hence, by Proposition \ref{thm:main2} and assumptions on the step sizes, it is straightforward to show that
\begin{align}
\E[V^*-V^{\pii_{\hat{T}}} ] 
\leq &\mathcal{O}\left(\frac{1}{T^{1-\sigma}}\right)+\begin{cases}
\tilde{\mathcal{O}}\left(\frac{1}{T^{\sigma}}\right) & \text{if}\quad 1> 2\sigma \\
\tilde{\mathcal{O}}(\frac{1}{T^{1-\sigma}}) & \text{o.w} 
\end{cases}+\begin{cases}
\tilde{\mathcal{O}}\left(\frac{\epsilon}{T^{1-\sigma}}\right) & \text{if}\quad \xi+\sigma > 1\\
\tilde{\mathcal{O}}(\frac{\epsilon}{T^{\xi}}) & \text{o.w} 
\end{cases}\nonumber\\
&+\frac{1}{T^{1-\sigma}}\mathcal{O}\left\{\sum_{t=0}^T  \beta_t\E\|Q^{\hat{\pi}_t}-Q_{t+1}\|\right\}.\label{eq:first_bound}
\end{align}
Next, we aim at bounding the term $\sum_{t=0}^T \beta_t\E\|Q^{\hat{\pi}_t}-Q_{t+1}\|$. We have:
\begin{align}
    \sum_{t=0}^T \beta_t\E\|Q^{\hat{\pi}_t}-Q_{t+1}\|=&   \sum_{t=0}^T \beta_t^{1/2\sigma}\beta_t^{1-1/2\sigma}\E\|Q^{\hat{\pi}_t}-Q_{t+1}\| \nonumber\\
    \leq &\sqrt{\sum_{t=0}^T\beta_t^{1/\sigma}}\times\sqrt{\sum_{t=0}^T\beta_t^{\frac{2\sigma-1}{\sigma}}\E\|Q^{\hat{\pi}_t}-Q_{t+1}\|^2},\label{eq:after_cauchy}
\end{align}
where the inequality is due to Cauchy–Schwarz inequality. Furthermore, we have $\sum_{t=0}^T\beta_t^{1/\sigma} = \mathcal{O}(\log(T))=\tilde{\mathcal{O}}(1)$. Hence, we only left to bound the last term in \eqref{eq:after_cauchy} which we will do in the rest of the proof.

Using $\|\theta_t\|^2$ as the Lyapunov function, we have:
\begin{align}
    \|\theta_{t+1}\|^2-\|\theta_t\|^2 =& 2\theta_t^\top(\theta_{t+1}-\theta_t-\alpha_t\Bar{A}^{\pii_{t-1}}\theta_t) + \|\theta_{t+1}-\theta_t\|^2 + 2\alpha_t\theta_t^\top\Bar{A}^{\pii_{t-1}}\theta_t\nonumber\\
    \stackrel{(a)}{=}&2\alpha_t\Gamma(\pii_{t-1},\theta_t,O_t) + 2\theta_t^\top(Q^{\pii_{t-1}}-Q^{\pii_{t}})+\|\theta_{t+1}-\theta_t\|^2+2\alpha_t\theta_t^\top\Bar{A}^{\pii_{t-1}}\theta_t\nonumber\\
    \stackrel{(b)}{\leq} & 2\alpha_t\Gamma(\pii_{t-1},\theta_t,O_t) + 2\|\theta_t\|.\overbrace{\|Q^{\pii_{t-1}}-Q^{\pii_{t}}\|}^{T_1}+\overbrace{\|\theta_{t+1}-\theta_t\|^2}^{T_2}+2\alpha_t\overbrace{\theta_t^\top\Bar{A}^{\pii_{t-1}}\theta_t}^{T_3},\label{eq:before_lemmas}
\end{align}
where $(a)$ is by definition of $\Gamma$, and $(b)$ is due to the Cauchy–Schwarz inequality. We bound each of the terms $T_1$, $T_2$, $T_3$ using Lemmas \ref{lem:cont3}, \ref{lem:D_dif_square3}, and \ref{lem:mu3}, respectively. We have
\begin{align}
\|\theta_{t+1}\|^2-\|\theta_t\|^2
    \leq &
    2\alpha_t\Gamma(\pii_{t-1},\theta_t,O_t) + 2\CLL\left(\CLLL\frac{\epsilon_{t-2}}{t-1}+\CL\beta_{t-1}\right)\|\theta_t\|+2\alpha_t^2\Delta_Q^2+4\CLL^2\CLLL^2\frac{\epsilon_{t-2}^2}{(t-1)^2}\nonumber\\
    &+4\CLL^2\CL^2\beta_{t-1}^2-\frac{2(1-\gamma)\mu}{|\mathcal{A}|}\alpha_t\epsilon_{t-2}\|\theta_t\|^2.\label{eq:before_sum}
\end{align}

Define $\lambda_t=\beta_t^{\frac{2\sigma-1}{\sigma}}$. Multiplying both sides of \eqref{eq:before_sum} with $\lambda_t$ and denoting $y_t=\lambda_t\|\theta_t\|^2$, we have $y_t \leq e_t(\|\theta_t\|^2-\|\theta_{t+1}\|^2)+u_t+h_t \sqrt{y_t}$, where $e_t=\frac{\lambda_t|\mathcal{A}|}{2(1-\gamma)\mu\alpha_t\epsilon_{t-2}}$ and $u_t = \frac{|\mathcal{A}|\lambda_t}{2(1-\gamma)\mu\alpha_t\epsilon_{t-2}}(2\alpha_t\Gamma(\pii_{t-1},\theta_t,O_t) +2\alpha_t^2\Delta_Q^2+4\CLL^2\CLLL^2\frac{\epsilon_{t-2}^2}{(t-1)^2}+4\CLL^2\CL^2\beta_{t-1}^2)$, and $h_t=\frac{|\mathcal{A}|\sqrt{\lambda_t}}{2(1-\gamma)\mu\alpha_t\epsilon_{t-2}}2\CLL\left(\CLLL\frac{\epsilon_{t-2}}{t-1}+\CL\beta_{t-1}\right)$. Summing from $\tau_t+2$ to $t$, we have
\begin{align}
\sum_{k=\tau_t+2}^t y_k \leq \underbrace{\sum_{k=\tau_t+2}^t e_k (\|\theta_k\|^2-\|\theta_{k+1}\|^2)}_{T_1} + \underbrace{u_k}_{T_2} + \underbrace{h_k\sqrt{y_k}
}_{T_3}.\label{eq:T1T2T3}
\end{align}

We bound the summation of each of the terms $T_1$, $T_2$, and $T_3$ separately. First we have
\begin{align*}
    T_1 =& e_{\tau_t+1}\|\theta_{\tau_t+2}\|^2 - e_{t}\|\theta_{t+1}\|^2  + \sum_{k=\tau_t+2}^t (e_k-e_{k-1}) \|\theta_k\|^2.
\end{align*}
We have $e_k \sim \lambda_k/(\alpha_k\epsilon_k)\sim \beta_k^{\frac{2\sigma-1}{\sigma}}/\alpha_k\epsilon_k\sim  k^{\nu+\xi+1-2\sigma}/\epsilon$. For the case $\nu+\xi+1-2\sigma > 0$, $e_k$ is increasing. Hence, we have 
\begin{align*}
    T_1&\stackrel{(a)}{\leq}
    4|\mathcal{S}||\mathcal{A}|Q_{\max}^2\left[e_{\tau_t+1} + \sum_{k=\tau_t+2}^t (e_k-e_{k-1})\right]\\
    &\stackrel{(b)}{\leq} 4 |\mathcal{S}||\mathcal{A}|Q_{\max}^2\left[e_{\tau_t+1} + e_t\right]\leq \tilde{\mathcal{O}}(t^{\nu+\xi+1-2\sigma}/\epsilon),
\end{align*}
where $(a)$ is due to Lemma \ref{lem:bounds3}, and $(b)$ is due to $e_t\geq0$. Furthermore, if $\nu+\xi+1-2\sigma < 0$, we have $e_k$ decreasing, and hence $T_1\leq \tilde{\mathcal{O}}(e_{\tau_t+1})=\tilde{\mathcal{O}}(1/\epsilon)$. It is also easy to show that for $\nu+\xi+1-2\sigma = 0$, we have $T_1\leq \tilde{\mathcal{O}}(1/\epsilon)$. Hence, in total we have 
\begin{align}
    T_1\leq\begin{cases}\tilde{\mathcal{O}}(t^{\nu+\xi+1-2\sigma}/\epsilon)& \text{if}~~ \nu+\xi+1>2\sigma,\\
\tilde{\mathcal{O}}(1/\epsilon)& \text{o.w.}\end{cases}\label{eq:T_one}
\end{align}
Furthermore, for the term $T_2$ we have
\begin{align}
\E T_2 =& \mathcal{O}(\sum_{k=\tau_t+2}^t\frac{\lambda_k}{\epsilon_k}\E \Gamma(\pii_{k-1},\theta_k,O_k)+\frac{\lambda_k\alpha_k}{\epsilon_k}+\frac{\lambda_k\epsilon_{k}}{k^2\alpha_k}+\frac{\lambda_k\beta_k^2}{\alpha_k\epsilon_k})\nonumber\\
\stackrel{(a)}{\leq}&\tilde{\mathcal{O}}(\sum_{k=\tau_t+2}^t\frac{\lambda_k}{\epsilon_k}(\beta_k+\alpha_k+\frac{\epsilon_k}{k})+\frac{\lambda_k\alpha_k}{\epsilon_k}+\frac{\lambda_k\epsilon_{k}}{k^2\alpha_k}+\frac{\lambda_k\beta_k^2}{\alpha_k\epsilon_k})\nonumber\\
\stackrel{(b)}{\leq}&\tilde{\mathcal{O}}(\sum_{k=\tau_t+2}^t k^{1-2\sigma}(k^{\xi-\nu}/\epsilon+k^{-1}+k^{\xi+\nu-2\sigma}/\epsilon))\nonumber\\
\leq& \begin{cases}\tilde{\mathcal{O}}(t^{2+\xi-2\sigma-\nu}/\epsilon)& \text{if}~~ 2+\xi>\nu+2\sigma,\\
\tilde{\mathcal{O}}(1/\epsilon)& \text{o.w.}\end{cases}+ \begin{cases}\tilde{\mathcal{O}}(t^{1-2\sigma})& \text{if}~~ 1>2\sigma,\\
\tilde{\mathcal{O}}(1)& \text{o.w.}\end{cases}\nonumber\\
&+\begin{cases}\tilde{\mathcal{O}}(t^{2+\xi+\nu-4\sigma}/\epsilon)& \text{if}~~ 2+\xi+\nu>4\sigma,\\
\tilde{\mathcal{O}}(1/\epsilon)& \text{o.w,}\end{cases}\label{eq:T_two}
\end{align}
where in $(a)$ we use Lemma \ref{lem:b_t3} with $\tau=\tau_k$, and in $(b)$ we use the assumptions on the step sizes. 

Finally, for the term $T_3$ we have
\begin{align}
    \E T_3 \stackrel{(a)}{\leq} & \sqrt{\sum_{k=\tau_t+2}^t h_k^2}\times \E\left[\sqrt{\sum_{k=\tau_t+2}^t y_k}\right]\nonumber\\
    \stackrel{(b)}{\leq}&  \sqrt{\sum_{k=\tau_t+2}^t h_k^2}\times \sqrt{\sum_{k=\tau_t+2}^t \E y_k},\nonumber
\end{align}
where $(a)$ is by Cauchy–Schwarz inequality and $(b)$ is by concavity of square root and Jensen's inequality. Denoting $G(t)=\sum_{k=\tau_t+2}^t h_k^2$, we have
\begin{align}
    G(t) \leq & \tilde{\mathcal{O}}\left(\sum_{k=\tau_t+2}^t\frac{\lambda_k}{\alpha_k^2k^2}+\frac{\lambda_k\beta_k^2}{\alpha_k^2\epsilon_k^2}\right)\nonumber\\
    \leq& \tilde{\mathcal{O}}\left(\sum_{k=\tau_t+2}^t k^{-1} +k^{1-4\sigma+2\nu+2\xi}/\epsilon^2\right)\nonumber\\
    =&\tilde{\mathcal{O}}(1) + \begin{cases}\tilde{\mathcal{O}}(t^{2-4\sigma+2\nu+2\xi}/\epsilon^2)& \text{if}~~ 2+2\nu+2\xi>4\sigma,\\
\tilde{\mathcal{O}}(1/\epsilon^2)& \text{o.w.}\end{cases}\label{eq:G_t_bound2}
\end{align}
Denote $H(t)=\sum_{k=\tau_t+2}^t\E y_k$. Taking expectation on both sides of \eqref{eq:T1T2T3}, we have 
\begin{align}
    H(t)&\leq \E T_1+\E T_2+ \sqrt{G(t)}.\sqrt{H(t)}\nonumber\\
    \implies (\sqrt{H(t)}-\frac{1}{2}\sqrt{G(t)})^2&\leq \E T_1+\E T_2 + 1/4 G(t)\nonumber\\
    \implies \sqrt{H(t)}-\frac{1}{2}\sqrt{G(t)}&\leq \sqrt{\E T_1+\E T_2 + 1/4 G(t)}\nonumber\\
    \implies H(t)&\leq 2\E T_1+2\E T_2 + G(t)\label{eq:H_t_bound}
\end{align}

Combining \eqref{eq:T_one}, \eqref{eq:T_two}, \eqref{eq:G_t_bound2}, and \eqref{eq:H_t_bound}, we have
\begin{align*}
    H(t) \leq&\begin{cases}\tilde{\mathcal{O}}(t^{\nu+\xi+1-2\sigma}/\epsilon)& \text{if}~~ \nu+\xi+1>2\sigma,\\
\tilde{\mathcal{O}}(1/\epsilon)& \text{o.w.}\end{cases}+\begin{cases}\tilde{\mathcal{O}}(t^{2+\xi-2\sigma-\nu}/\epsilon)& \text{if}~~ 2+\xi>\nu+2\sigma,\\
\tilde{\mathcal{O}}(1/\epsilon)& \text{o.w.}\end{cases}\nonumber\\&+ \begin{cases}\tilde{\mathcal{O}}(t^{1-2\sigma})& \text{if}~~ 1>2\sigma,\\
\tilde{\mathcal{O}}(1)& \text{o.w.}\end{cases}+\begin{cases}\tilde{\mathcal{O}}(t^{2+\xi+\nu-4\sigma}/\epsilon)& \text{if}~~ 2+\xi+\nu>4\sigma,\\
\tilde{\mathcal{O}}(1/\epsilon)& \text{o.w,}\end{cases}\\
& + \begin{cases}\tilde{\mathcal{O}}(t^{2-4\sigma+2\nu+2\xi}/\epsilon^2)& \text{if}~~ 2+2\nu+2\xi>4\sigma,\\
\tilde{\mathcal{O}}(1/\epsilon^2)& \text{o.w.}\end{cases}+\tilde{\mathcal{O}}(1).
\end{align*}
Combining the above bound, \eqref{eq:first_bound} and \eqref{eq:after_cauchy}, we get the result.\QED
\subsubsection{Proof of Corollary \ref{cor:1.1}}

In the case of constant exploration parameter, we have $\xi=0$, and the optimal step size can be achieved by $\sigma=3/4$ and $\nu=1/2$. In this case, we get $\E[V^*-V^{\pii_{\hat{T}}}]\leq\tilde{\mathcal{O}}\left(\frac{T^{-1/4}}{\epsilon} +\epsilon\right)$. Hence, to get to a solution policy within $\delta/\epsilon+\epsilon$ of the global optimum, we need $\tilde{\mathcal{O}}(1/\delta^4)$ number of samples. Furthermore, to get $\delta$-close to the global optimum, we should have $\tilde{\mathcal{O}}(\frac{T^{-1/4}}{\epsilon})\leq \delta/2$ and $\tilde{\mathcal{O}}(\epsilon)\leq \delta/2$, which means we have $\tilde{\mathcal{O}}(T^{-1/8})\leq \delta$. Hence, to get $\delta$-close to the global optimum, we need $\tilde{\mathcal{O}}(1/\delta^{8})$ number of samples. 
\QED
\subsubsection{Proof of Corollary \ref{cor:1.2}}
For $\xi>0$ we get $\E[V^*-V^{\pii_{\hat{T}}}]\leq\tilde{\mathcal{O}}(T^{-1/6})$ convergence to the global optimum which can be achieved by $\xi=1/6$, $\nu=1/2$, and $\sigma=5/6$. Hence, in this case to get $\delta$-close to the global optimum, we need $\tilde{\mathcal{O}}(1/\delta^{6})$ number of samples. This proves Corollaries \ref{cor:1.1} and \ref{cor:1.2}.
\QED

%% file: conclusion.tex
\section{Conclusion}
In this paper we studied the convergence of two-time-scale natural actor-critic algorithm. In order to promote exploration and ensure convergence, we employed $\epsilon$-greedy in the iterations of the algorithm. We have shown that with a constant $\epsilon$ parameter, the actor-critic algorithm converges to a ball
around the global optimum with radius $\epsilon+\delta/\epsilon$ using $\tilde{\mathcal{O}}(1/\delta^{4})$ number of samples, and using a small enough exploration parameter $\epsilon$ it requires $\tilde{\mathcal{O}}(1/\delta^{8})$ number of samples to find a policy within $\delta$-ball around the global optimum. Furthermore, with a carefully diminishing greedy parameter, we show $\Tilde{\mathcal{O}}(\delta^{-6})$ sample complexity for the convergence to the global optimum. Due to the employment of the $\epsilon$-greedy, we characterize this sample complexity with the minimum set of assumptions, i.e. the ergodicity of the underlying MDP. We show that this assumption is indeed necessary for establishing the convergence of natural actor-critic algorithm. Improving this sample complexity, and characterizing the same sample complexity in the function approximation setting is among our future works.  

%% file: best_thm_proof.tex
\section{Details of the Proof of Theorem \ref{thm: improv_mix}} \label{sec:C}

\subsection{Proof of Useful Lemmas}
\textit{Proof of Lemma \ref{lem:D_dif_square3}:}
\begin{align*}
    \|\theta_{t+1}-\theta_t\|^2 &= \|Q_{t+1}-Q_t+Q^{\pii_{t-1}}-Q^{\pii_t}\|^2\\
    &\leq 2\|Q_{t+1}-Q_t\|^2 + 2\|Q^{\pii_{t-1}}-Q^{\pii_t}\|^2\\
    &\stackrel{(a)}{\leq} 2\alpha_t^2\Delta_Q^2+2\CLL^2\|\pii_{t-1}-\pii_t\|^2\\
    &\stackrel{(b)}{\leq} 2\alpha_t^2\Delta_Q^2+ 2\CLL^2\left(\CLLL \frac{\epsilon_{t-2}}{t-1} + \CL \beta_{t-1}\right)^2\\
    &\leq 2\alpha_t^2\Delta_Q^2+4\CLL^2\CLLL^2\frac{\epsilon_{t-2}^2}{(t-1)^2}+4\CLL^2\CL^2\beta_{t-1}^2,
\end{align*}
where $(a)$ is due to Lemmas \ref{lem:cont4} and \ref{lem:bounds}, and $(b)$ is due to Lemma \ref{lem:pi_lip3}.
\QEDopen

\textit{Proof of Lemma \ref{lem:mu3}:}
We prove this lemma for a slightly more general case. Assume a finite state Markov chain $\{X_k\}_{k=0,1,\dots}$ with state space $\mathcal{X}=\{x_1,x_2,\dots,x_{|\mathcal{X}|}\}$  and stationary distribution $\nu$. Define $M:=diag(\nu)$ a diagonal matrix with diagonal entries equal to the elements of $\nu$. Clearly, $M=M^\top$. Further denote $P$ as the transition matrix of the Markov chain. Define $V=\gamma P-I$, where $I$ is the identity matrix. Assuming $X_k\sim \nu$, for any function $F(\cdot):\mathcal{X}\rightarrow \mathbb{R}$, we have:
\[
\E\left[F\left(X_k\right)^2\right]=\E\left[F\left(X_{k+1}\right)^2\right].
\]
By Cauchy–Schwarz inequality, we have:
\begin{align}
\E\left[F\left(X_k\right)F\left(X_{k+1}\right)\right]&\leq \sqrt{\E\left[F^2(X_k)\right]}.\sqrt{\E\left[F^2(X_{k+1})\right]} \nonumber\\
&= \E\left[F^2(X_k)\right].\label{eq:cauchy}
\end{align}
Denoting $F=[F(x_1);F(x_2);\dots;F(x_{|\mathcal{X}|})]$ as a $|\mathcal{X}|$ dimentional vector, we have:
\begin{align}
\E[F^2(X_k)]&=\sum_{x\in\mathcal{X}}\nu(x)F^2(x)=F^\top M F,\label{eq:FM}\\ \E[F(X_k)F(X_{k+1})]&=\sum_{x,y\in\mathcal{X}}\nu(x)P(y|x)F(x)F(y) \nonumber\\
&= F^\top M P F=F^\top P^\top M F,\label{eq:FM2}
\end{align}
where the last equality is due to $\E[F(X_k)F(X_{k+1})]$ being a scalar. Combining \eqref{eq:cauchy}, \eqref{eq:FM}, and \eqref{eq:FM2}, we have:
\begin{align}
F^\top M P F &\leq F^\top M F, ~ ~\forall F\nonumber\\
\implies M P&\leq M\nonumber\\
\implies M(\gamma P-I)&\leq -(1-\gamma) M.\label{eq:final_ineq}
\end{align}

Next, in the case of MDP, for a fixed policy $\pi$, we define $M^\pi\in \mathbb{R}^{|\mathcal{S}||\mathcal{A}|\times|\mathcal{S}||\mathcal{A}|}$ and $P^\pi\in \mathbb{R}^{|\mathcal{S}||\mathcal{A}|\times|\mathcal{S}||\mathcal{A}|}$ matrices as follows:
\begin{align*}
M^\pi_{(s,a),(s',a')}&=\begin{cases}\mu^\pi(s)\pi(a|s)& (s,a)=(s',a'),\\
0& o.w\end{cases}\\
P^\pi_{(s,a),(s',a')}&=\mathcal{P}(s'|s,a)\pi(a'|s').
\end{align*}
It is easy to see that:
\begin{align*}
\bar{A}^\pi_{(s,a),(s',a')}
&=\begin{cases}\mu^\pi(s)\pi(a|s)\left(\gamma \mathcal{P}\left(s'|s,a\right)\pi\left(a'|s'\right)-1\right) & s=s', a=a',\\
\gamma\mu^\pi(s)\pi(a|s) \mathcal{P}\left(s'|s,a\right)\pi\left(a'|s'\right) & s\neq s'~ \text{or}~ a\neq a'
\end{cases}\\
\implies \bar{A}^\pi&=M^\pi(\gamma P^\pi-I)\leq -(1-\gamma)M^\pi,
\end{align*}
where the last inequality follows from \eqref{eq:final_ineq}. As a result, we have:
\begin{align*}
\theta^\top_t \bar{A}^{\pii_{t-1}} \theta_t &\leq -(1-\gamma) \sum_{s,a} \mu^{\pi}(a)\pii_{t-1}(a|s)\theta_{t,s,a}^2\\
&\leq -(1-\gamma)\frac{\epsilon_{t-2}}{|\mathcal{A}|}\mu||\theta_t||^2,    
\end{align*}
where the last inequality follows from $\pii_{t-1}(a|s)\geq \frac{\epsilon_{t-2}}{|\mathcal{A}|}$ and Lemma \ref{lem:mulem2}.\QEDopen

\textit{Proof of Lemma \ref{lem:bounds3}:}
\begin{enumerate}
\item By the assumption on the reward function $\mathcal{R}(s,a)\geq 0$, we have $V^{\pi}(s)=\mathbb{E}\left[\sum_{k=0}^{\infty} \gamma^{k}\mathcal{R}(S_{k},A_{k}) \,|\,S_{0} = s\right]\geq 0$. Furthermore, due to $\mathcal{R}(s,a)\leq 1$, we have $V^{\pi}(s)=\mathbb{E}\left[\sum_{k=0}^{\infty} \gamma^{k}\mathcal{R}(S_{k},A_{k}) \,|\,S_{0} = s\right]\leq \mathbb{E}\left[\sum_{k=0}^{\infty} \gamma^{k} \,|\,S_{0} = s\right] = \frac{1}{1-\gamma}$ for all $s\in\mathcal{S}$.
\item Similarly, we have $Q^\pi(s,a)\in[0,\frac{1}{1-\gamma}]$ for all $s,a$. 
\item
$\|Q^\pi\| =\sqrt{\sum_{s,a}{Q^\pi}^2(s,a)}\leq \frac{\sqrt{|\mathcal{S}||\mathcal{A}|}}{1-\gamma}$
\item In order to prove this, first we show $\|Q_t\|_\infty\leq \frac{1}{1-\gamma}$ for all $t\geq0$. We construct this bound by induction. Due to the initialization, the inequality holds for $t=0$. Assuming the inequality holds for $t$, we prove it holds for $t+1$. For all $s,a$, we have:
    \begin{align*}
    |Q_{t+1}(s,a)|
    =&\left|(1-\alpha_t(s,a))Q_{t}(s,a)+\alpha_t(s,a)(\mathcal{R}(s,a)+\gamma Q_{t}(s_{t+1},a_{t+1}))\right|\\
    \leq& (1-\alpha_t(s,a))|Q_{t}(s,a)|+\alpha_t(s,a)|\mathcal{R}(s,a)+\gamma Q_{t}(S_{t+1},A_{t+1})|\\
    \leq& (1-\alpha_t(s,a))Q_{\max} + \alpha_t(s,a) (1+\gamma Q_{\max})\\
    =&(1-\alpha_t(s,a))\frac{1}{1-\gamma} +\alpha_t(s,a) (1+\gamma \frac{1}{1-\gamma}) 
    = \frac{1}{1-\gamma}.
    \end{align*}
    The bound for $\|Q_t\|$ follows directly. \QEDopen
\end{enumerate}

\textit{Proof of Lemma \ref{lem:b_t3}:}
Given time indices $t$ and $\tau<t$, we consider the following auxiliary chain of state-actions:
\begin{align*}
S_{t-\tau}&\xrightarrow{\pii_{t-\tau-1}} A_{t-\tau}\xrightarrow{\mathcal{P}}\Tilde{S}_{t-\tau+1}\xrightarrow{\pii_{t-\tau-1}} \Tilde{A}_{t-\tau+1}\dots\xrightarrow{\mathcal{P}}\Tilde{S}_t\xrightarrow{\pii_{t-\tau-1}}\Tilde{A}_t\xrightarrow{\mathcal{P}}\Tilde{S}_{t+1}\xrightarrow{\pii_{t-\tau-1}}\Tilde{A}_{t+1}.
\end{align*}
Note that the original chain is as follows:
\begin{align*}
S_{t-\tau}&\xrightarrow{\pii_{t-\tau-1}} A_{t-\tau}\xrightarrow{\mathcal{P}}S_{t-\tau+1}\xrightarrow{\pii_{t-\tau}} A_{t-\tau+1}\dots\xrightarrow{\mathcal{P}}S_t\xrightarrow{\pii_{t-1}}A_t\xrightarrow{\mathcal{P}}S_{t+1}\xrightarrow{\pii_{t}}A_{t+1}.
\end{align*}
Further, we define $\tilde{O}_t=(\Tilde{S}_t,\Tilde{A}_t,\Tilde{S}_{t+1},\Tilde{A}_{t+1})$. We have:
\begin{align}
    \Gamma(\pii_{t-1},\theta_t,O_t)=&\Gamma(\pii_{t-1},\theta_t,O_t) - \Gamma(\pii_{t-\tau-1},\theta_t,O_t)\\
    &+\Gamma(\pii_{t-\tau-1},\theta_t,O_t)-\Gamma(\pii_{t-\tau-1},\theta_{t-\tau},O_t)\\
    &+\Gamma(\pii_{t-\tau-1},\theta_{t-\tau},O_t)-\Gamma(\pii_{t-\tau-1},\theta_{t-\tau},\tilde{O}_t)\\
    &+\Gamma(\pii_{t-\tau-1},\theta_{t-\tau},\tilde{O}_t).
\end{align}

We bound each of the terms above separately. Firstly:
\begin{align*}
\Gamma(\pii_{t-1},\theta_t,O_t) -\Gamma(\pii_{t-\tau-1},\theta_t,O_t) \stackrel{(a)}{\leq}& K_1\|\pii_{t-1}-\pii_{t-\tau-1}\|\\
\stackrel{(b)}{\leq}& K_1\sum_{i=t-\tau}^{t-1}\left\|\pii_i-\pii_{i-1}\right\|\\
\stackrel{(c)}{\leq}& K_1\sum_{i=t-\tau}^{t-1} \left[\CLLL\frac{\epsilon_{i-2}}{i-1}+\CL\beta_{i-1}\right]\\
\leq& K_1\tau\left[\CLLL\frac{\epsilon_{t-\tau-2}}{t-\tau-1}+\CL\beta_{t-\tau-1}\right],
\end{align*}
where $(a)$ is due to Lemma \ref{lem:D.4}, $(b)$ is by triangle inequality, and $(c)$ is due to Lemma \ref{lem:pi_lip3}. Second, we have:
\begin{align*}
\Gamma(\pii_{t-\tau-1},\theta_t,O_t) -\Gamma(\pii_{t-\tau-1},\theta_{t-\tau},O_t)\stackrel{(a)}{\leq}& K_2\|\theta_t-\theta_{t-\tau}\|\\
\stackrel{(b)}{\leq}& K_2\sum_{i=t-\tau+1}^t\|\theta_i-\theta_{i-1}\|\\
 \stackrel{(c)}{\leq}& K_2 \sum_{i=t-\tau+1}^t \Delta_Q\alpha_{i-1}+\CLL \CLLL\frac{\epsilon_{i-3}}{i-2} +  \CL \CLL\beta_{i-2}\\
\leq& K_2 \tau \left[\Delta_Q\alpha_{t-\tau}+\CLL \CLLL\frac{\epsilon_{t-\tau-2}}{t-\tau-1} +  \CL \CLL\beta_{t-\tau-1}\right],
\end{align*}
where $(a)$ due to Lemma \ref{lem:D.5}, $(b)$ is by triangle inequality, and $(c)$ is due to Lemma \ref{lem:bounds}. Third, denoting $\mathcal{F}_{t-\tau}:=\{S_{t-\tau},\pii_{t-\tau-1},\theta_{t-\tau}\}$, we have:
\begin{align*}
\E \bigg[\Gamma(\pii_{t-\tau-1},\theta_{t-\tau},O_t) -\Gamma(\pii_{t-\tau-1},\theta_{t-\tau},\tilde{O}_t)\bigg|\mathcal{F}_{t-\tau}\bigg]\stackrel{(a)}{\leq}& C_u\E \left[\sum_{i=t-\tau}^t\|\pii_i-\pii_{t-\tau-1}\|\Bigg|\mathcal{F}_{t-\tau}\right] \\
\stackrel{(b)}{\leq}& C_u(\tau+1)^2\left[\CLLL\frac{\epsilon_{t-\tau-2}}{t-\tau-1}+\CL\beta_{t-\tau-1}\right],
\end{align*}
where $(a)$ is due to Lemma \ref{lem:D.6} and $(b)$ is due to Lemma \ref{lem:my_lemma2}. Finally, by Lemma \ref{lem:D.7} we have:
\[
\E \left[\Gamma(\pii_{t-\tau-1},\theta_{t-\tau},\Tilde{O}_t)\big|\mathcal{F}_{t-\tau}\right] \leq C_bm\rho^{\tau}.
\]
Combining the bounds above, and noticing $\tau\geq 1$, we get the result.\QEDopen

\textit{Proof of Lemma \ref{lem:mixing3}:}
Suppose $\pi$ be an arbitrary stochastic policy. $\pi$ can be written as a $|\mathcal{S}|$ by $|\mathcal{A}|$ stochastic matrix, which has non-negative elements, and each row sums up to one. Hence, by \cite[Theorem 1]{davis1961markov}, $\pi$ can be written as a convex combination of at most $N=|\mathcal{S}|(|\mathcal{A}|-1)+1$ deterministic policies $\{\pi_i\}_{i=1}^{|\mathcal{S}|(|\mathcal{A}|-1)+1}$. In other words, there exist coefficients $\{\alpha_i\}_{i=1}^{N}$, such that $\alpha_i\geq 0$ and $\sum_i \alpha_i=1$, and $\pi=\sum_i \alpha_i\pi_i$. By definition of $P^\pi$, we have $P^\pi=\sum_i\alpha_iP^{\pi_i}$. 

Due to ergodicity Assumption \ref{ass:stat_pos}, for every policy $\pi_i$, there exist a finite integer $r_i$, such that $(P^{\pi_i})^{\bar{r}_i}$ is a positive matrix with minimum element $e_i>0$ for all $\bar{r}_i\geq r_i$. Since we have a finite number of $r_i$ and $e_i$'s, we have $r=\max_i r_i$ is a finite integer, and $e = \min_i e_i > 0$. Furthermore, we have \begin{align}
    \left[(P^\pi)^r\right]_{m,n} =& \left[\left(\sum_i\alpha_iP^{\pi_i}\right)^r\right]_{m,n}\\
    \stackrel{(a)}{\geq}& \sum_i\alpha_i^r \left[(P^{\pi_i})^r\right]_{m,n}\geq N e \frac{1}{N}\sum_i \alpha_i^r\\ \stackrel{(b)}{\geq}& Ne \left(\frac{1}{N}\sum_i\alpha_i\right)^r = Ne\left(\frac{1}{N}\right)^r > 0,\nonumber
\end{align}
where $(a)$ is due to non-negativity of matrices $\alpha_iP^{\pi_i}$ and $(b)$ is by Jensen's inequality. Hence, by \cite[Theorem 4.9]{levin2017markov}, we can show the existence of $\rho\in(0,1)$ and $m>0$.

Furthermore, if the underlying Markov chain under a policy $\pi$ is periodic with period $d$, then we have $\lim_{t\rightarrow\infty} P(S_{dt}=i|S_0=i) > 0 $ while $\lim_{t\rightarrow\infty}P(S_{dt+1}=i|S_0=i)=0$, and hence \eqref{eq:mixing3} does not hold.\QEDopen

\textit{Proof of Lemma \ref{lem:cont4}:}
By the policy gradient theorem \cite{agarwal2019theory}, we know that for any distribution $\mu$, we have $\frac{\partial V^\pi(\mu)}{\partial \pi(a|s)}=\frac{1}{1-\gamma}d_\mu^\pi(s)Q^\pi(s,a)$. As a result:
\begin{align*}
\left\|\frac{\partial V^\pi(\mu)}{\partial \pi}\right\| &\leq \frac{1}{1-\gamma}\sqrt{\sum_{s,a}{Q^\pi}^2(s,a)} \leq \frac{\sqrt{|\mathcal{S}||\mathcal{A}|}}{(1-\gamma)^2}.     
\end{align*}
Furthermore, we have:
\begin{align*}
\frac{\partial Q^{\pi}(s,a)}{\partial\pi}&=\gamma\sum_{s'}\mathcal{P}(s'|s,a)\frac{\partial V^{\pi}(s')}{\partial\pi},
\end{align*}
which implies
\begin{align*}
 \left\|\frac{\partial Q^{\pi}(s,a)}{\partial\pi}\right\| &\leq \gamma\sum_{s'}\mathcal{P}(s'|s,a)\left\|\frac{\partial V^{\pi}(s')}{\partial\pi}\right\|\leq \gamma\frac{\sqrt{|\mathcal{S}||\mathcal{A}|}}{(1-\gamma)^2} \\
\implies|Q^{\pi_1}(s&,a)-Q^{\pi_2}(s,a)| \leq \gamma\frac{\sqrt{|\mathcal{S}||\mathcal{A}|}}{(1-\gamma)^2} \|\pi_1-\pi_2\|.
\end{align*}

Using this, we have:
\begin{align*}
    \|Q^{\pi_1}-Q^{\pi_2}\| &\leq \sqrt{\sum_{s,a} \frac{\gamma^2|\mathcal{S}||\mathcal{A}|}{(1-\gamma)^4} \|\pi_1-\pi_2\|^2}\\
    &=  \frac{\gamma|\mathcal{S}||\mathcal{A}|}{(1-\gamma)^2} \|\pi_1-\pi_2\| = \CLL\|\pi_1-\pi_2\|,
\end{align*}
where $\CLL:=\frac{\gamma|\mathcal{S}||\mathcal{A}|}{(1-\gamma)^2}$.
\QEDopen

\textit{Proof of Lemma \ref{lem:pi_lip3}:}
Policy $\pi_t$ can be parameterized by the vector $\theta^t\in\mathbb{R}^{|\mathcal{S}||\mathcal{A}|}$ as $\pi_t(a|s)=\frac{\exp(\theta^t_{s,a})}{\sum_{a'}\exp(\theta^t_{s,a'})}$. It is straightforward to see that the multiplicative weight update of the policy in Algorithm \ref{alg:1} is equivalent to \cite[Lemma 3.1]{chen2021finite}
\[
\theta^{t+1}=\theta^t+\beta_t Q_{t+1}.
\]
We have 
\begin{align}
\|\pi_{t+1}-\pi_t\|^2&=\sum_{s}\|\pi_{t+1}(\cdot|s)-\pi_{t}(\cdot|s)\|^2\nonumber\\
&\stackrel{(a)}{\leq} \sum_{s}\|\beta_t Q_{t+1}(s,\cdot)\|^2\leq \beta_t^2|\mathcal{A}||\mathcal{S}|Q_{\max}^2,\label{eq:pi_diff_bound}
\end{align}
where in $(a)$ we use $1$-Lipschitzness of the softmax function \cite{beck2017first}.

As a result, we have:
\begin{align*}
\|\pii_{t+1}-\pii_t\| 
&= \|(\epsilon_t-\epsilon_{t-1})(\frac{1}{\mathcal{A}}-\pi_{t+1})+(1-\epsilon_{t-1})(\pi_{t+1}-\pi_t)\| \\
& \stackrel{(a)}{\leq} |\epsilon_t-\epsilon_{t-1}|\sqrt{|\mathcal{S}|}(\frac{1}{\sqrt{|\mathcal{A}|}}+1) + \|\pi_{t+1}-\pi_t\|\\
&\stackrel{(b)}{\leq} \frac{\xi \epsilon_{t-1}}{t} \sqrt{|\mathcal{S}|}(\frac{1}{\sqrt{|\mathcal{A}|}}+1) + \beta_tQ_{\max}\sqrt{|\mathcal{A}||\mathcal{S}|}\\
&= \CLLL \frac{\epsilon_{t-1}}{t} + \CL \beta_t,
\end{align*}
where $(a)$ is due to triangle inequality and $(b)$ is due to the assumption on $\epsilon_t$ and \eqref{eq:pi_diff_bound}. Here $\CLLL=\xi\sqrt{|\mathcal{S}|}(\frac{1}{\sqrt{|\mathcal{A}|}}+1)$ and $\CL=Q_{\max}\sqrt{|\mathcal{A}||\mathcal{S}|}$.\QEDopen

\textit{Proof of Lemma \ref{lem:mulem2}:}
Ergodicity assumption \ref{ass:stat_pos} implies that the underlying Markov chain induced by all the policies is irreducible. The proof follows from \cite[Proposition 1.14]{levin2017markov}.
\QEDopen

\subsection{Auxiliary Lemmas}
\begin{lemma}\label{lem:D.4}
For any $\pi_1, \pi_2, \theta$, and $O=(S,A,S',A')$, 
\[
\left|\Gamma(\pi_1,\theta,O)-\Gamma(\pi_2,\theta,O)\right|\ \leq K_1 \|\pi_1-\pi_2\|,
\]
where $K_1=2Q_{\max}\sqrt{2|\mathcal{S}||\mathcal{A}|}\CLL+8Q_{\max}^2|\mathcal{S}|^2|\mathcal{A}|^3\left(\left\lceil\log_\rho m^{-1}\right\rceil+\frac{1}{1-\rho}+2\right)$.
\end{lemma}

\begin{lemma}\label{lem:D.5}
For any $\pi,  Q_1,Q_2$, and $O=(S,A,S',A')$, 
\[
|\Gamma(\pi,\theta_1,O)-\Gamma(\pi,\theta_2,O)| \leq K_2\|\theta_1-\theta_2\|,
\]
where $K_2=1+9\sqrt{2|\mathcal{S}||\mathcal{A}|}Q_{\max}$.
\end{lemma}
\begin{lemma}\label{lem:D.6}
Consider original tuples $O_t=(S_t,A_t,S_{t+1},A_{t+1})$ and the auxiliary tuples $\Tilde{O}_t=(\Tilde{S}_t,\Tilde{A}_t,\Tilde{S}_{t+1},\Tilde{A}_{t+1})$. Denote $\mathcal{F}_{t-\tau}:=\{S_{t-\tau},\pii_{t-\tau-1},\theta_{t-\tau}\}$. For any time indices $t>\tau>1$, we have
\begin{align*}
\E\left[\Gamma(\pii_{t-\tau-1},\theta_{t-\tau},O_t)-\Gamma(\pii_{t-\tau-1},\theta_{t-\tau},\tilde{O}_t)\mid\mathcal{F}_{t-\tau}\right] \leq  C_u\E\left[\sum_{i=t-\tau}^t\|\pii_i-\pii_{t-\tau-1}\|\mid\mathcal{F}_{t-\tau}\right], 
\end{align*}
where $C_u=4Q_{\max}|\mathcal{S}|^{1.5}|\mathcal{A}|^{1.5}(1+3Q_{\max}|\mathcal{S}||\mathcal{A}|)$.
\end{lemma}
\begin{lemma}\label{lem:D.7}
Consider the auxiliary tuple $\Tilde{O}_t=(\Tilde{S}_t,\Tilde{A}_t,\Tilde{S}_{t+1},\Tilde{A}_{t+1})$. Denote $\mathcal{F}_{t-\tau}=\{S_{t-\tau},\pii_{t-\tau-1},\theta_{t-\tau}\}$. For any time indices $t>\tau>1$, we have:
\[
\E\left[\Gamma(\pii_{t-\tau-1},\theta_{t-\tau},\Tilde{O}_t)\big|\mathcal{F}_{t-\tau}\right] \leq C_b m\rho^{\tau},
\]
where $C_b=4Q_{\max}|\mathcal{S}||\mathcal{A}|(1+3|\mathcal{S}||\mathcal{A}|Q_{\max})$.
\end{lemma}
\begin{lemma}\label{lem:my_lemma2}
For any time indices $t>\tau>1$, the policies generated by Algorithm \ref{alg:1} satisfy the following:
\[
\sum_{i=t-\tau}^t\left\|\pii_i-\pii_{t-\tau-1}\right\| \leq (\tau+1)^2\left[\CLLL\frac{\epsilon_{t-\tau-2}}{t-\tau-1}+\CL\beta_{t-\tau-1}\right].
\]
\end{lemma}
\begin{lemma}\label{lem:bounds}
We have the following bounds:
\begin{enumerate}
    \item $\|A(O)\|\leq \sqrt{1+\gamma^2} \leq \sqrt{2}$,
    \item $\|r(O)\|\leq 1$, 
    \item $\|\Bar{A}^\pi\|\leq \sqrt{2}$,
    \item $\left\| \E[r\left(O_1\right)-r(O_2)] \right\|_1\leq 2|\mathcal{S}||\mathcal{A}|d_{TV}(O_1,O_2)$,
    \item $\left\|\E[A\left(O_1\right)-A(O_2)]\right\|_1\leq 2|\mathcal{S}||\mathcal{A}|d_{TV}(O_1,O_2)$,
    \item $\|Q_{t+1}-Q_t\|\leq \alpha_t\Delta_{Q}:=\alpha_t(2Q_{\max}+1)$,
    \item $\left\|\theta_t-\theta_{t-1}\right\|\leq \Delta_Q\alpha_{t-1}+\CLL \CLLL\frac{\epsilon_{t-3}}{t-2} +  \CL \CLL\beta_{t-2}$,
\end{enumerate}
where $Q_{\max}=\frac{1}{1-\gamma}$, and the constants $\CL, \CLL, \CLLL$ are defined in Lemmas  \ref{lem:cont4} and \ref{lem:pi_lip3}.
\end{lemma}
\begin{lemma}\label{lem:C.22}
Consider $O_t=(S_t,A_t,S_{t+1},A_{t+1})$ and $\Tilde{O}_t=(\Tilde{S}_t,\Tilde{A}_t,\Tilde{S}_{t+1},\Tilde{A}_{t+1})$. Denote $\mathcal{F}_{t-\tau}:=\{S_{t-\tau},\pii_{t-\tau-1},\theta_{t-\tau}\}$. We have:
\begin{align*}
d_{TV}\big(P(O_t\in\cdot|&\mathcal{F}_{t-\tau})||P(\tilde{O}_t\in\cdot|\mathcal{F}_{t-\tau})\big)  \leq \sqrt{|\mathcal{A}||\mathcal{S}|}\E\left[\sum_{i=t-\tau}^t\|\pii_i-\pii_{t-\tau-1}\|\mid\mathcal{F}_{t-\tau}\right] 
\end{align*}
\end{lemma}

\begin{lemma}[Lemma A.1 in \cite{wu2020finite}]\label{lem:A.12}
Denote $M=\left\lceil\log_\rho m^{-1}\right\rceil+\frac{1}{1-\rho}$. For any $\pi_1$ and $\pi_2$ policies, we have the following inequality:
\begin{align*}
&d_{TV}\left(\mu^{\pi_1}\otimes\pi_1\otimes P\otimes\pi_1,\mu^{\pi_2}\otimes\pi_2\otimes P\otimes\pi_2\right)\leq |\mathcal{A}|\left(M+2\right)\|\pi_1-\pi_2\|
\end{align*}
\end{lemma}

\subsection{Proofs of the auxiliary Lemmas}
\textit{Proof of Lemma \ref{lem:D.4}:}
\begin{align*}
\Gamma(\pi_1,\theta,O)&-\Gamma(\pi_2,\theta,O) \\
=& \theta^\top A(O)(Q^{\pi_1}-Q^{\pi_{2}})-\theta^\top(\Bar{A}^{\pi_1}-\Bar{A}^{\pi_{2}})\theta\\
\stackrel{(a)}{\leq}& \|\theta\|.\|A(O)\|.\|Q^{\pi_1}-Q^{\pi_{2}}\| + \|\theta\|^2 . \|\Bar{A}^{\pi_1}-\Bar{A}^{\pi_{2}}\|\\
\stackrel{(b)}{\leq}& 2Q_{\max}\sqrt{2|\mathcal{S}||\mathcal{A}|}\|Q^{\pi_1}-Q^{\pi_{2}}\|+ 4Q_{\max}^2|\mathcal{S}||\mathcal{A}|.\|\Bar{A}^{\pi_1}-\Bar{A}^{\pi_{2}}\|\\
\stackrel{(c)}{\leq}&2Q_{\max}\sqrt{2|\mathcal{S}||\mathcal{A}|}\CLL\|\pi_1-\pi_{2}\|+8Q_{\max}^2|\mathcal{S}|^2|\mathcal{A}|^2 d_{TV}\left(\mu^{\pi_1}\otimes\pi_1\otimes \mathcal{P}\otimes\pi_1,\mu^{\pi_{2}}\otimes\pi_{2}\otimes \mathcal{P}\otimes\pi_{2}\right)\\
\stackrel{(d)}{\leq}&2Q_{\max}\sqrt{2|\mathcal{S}||\mathcal{A}|}\CLL\|\pi_1-\pi_{2}\|+8Q_{\max}^2|\mathcal{S}|^2|\mathcal{A}|^3\left(\left\lceil\log_\rho m^{-1}\right\rceil+\frac{1}{1-\rho}+2\right)\|\pi_1-\pi_{2}\|\\
=& K_1\|\pi_1-\pi_{2}\|
\end{align*}
where $(a)$ is due to Cauchy–Schwarz inequality, $(b)$ is due to Lemmas  \ref{lem:bounds3} and \ref{lem:bounds}, $(c)$ is due to Lemmas \ref{lem:cont4} and \ref{lem:bounds}, $(d)$ is due to Lemma \ref{lem:A.12}.\QEDopen

\textit{Proof of Lemma \ref{lem:D.5}:}
\begin{align*}
|\Gamma(\pi,\theta_1,O)-\Gamma(\pi,\theta_2,O)|
\stackrel{(a)}{\leq} &(\|r(O)\|+\|A(O)\|.\|Q^\pi\|)\|\theta_1-\theta_2\|+\|A(O)-\Bar{A}^\pi\|.\|\theta_1-\theta_2\|(\|\theta_1\|+\|\theta_{2}\|)\\
\stackrel{(b)}{\leq}&(1+9\sqrt{2|\mathcal{S}||\mathcal{A}|}Q_{\max})\|\theta_1-\theta_2\|
\end{align*}
where $(a)$ follows from Cauchy–Schwarz and triangle inequality, and $(b)$ is due to Lemmas \ref{lem:bounds3} and \ref{lem:bounds}.
\QEDopen

\textit{Proof of Lemma \ref{lem:D.6}:}
We have:
\begin{align*}
&\E\big[\Gamma(\pii_{t-\tau-1},\theta_{t-\tau},O_t)-\Gamma(\pii_{t-\tau-1},\theta_{t-\tau},\tilde{O}_t)|\mathcal{F}_{t-\tau}\big]\\
=&\theta_{t-\tau}^\top \E\left[r\left(O_t\right)-r(\tilde{O}_t)+\left(A\left(O_t\right)-A(\tilde{O}_t)\right)Q^{\pii_{t-\tau-1}}|\mathcal{F}_{t-\tau}\right]+\theta_{t-\tau}^\top\E\left[A(O_t)-A(\tilde{O}_t)|\mathcal{F}_{t-\tau}\right]\theta_{t-\tau}\\
\stackrel{(a)}{\leq} &\|\theta_{t-\tau}\|_\infty\left\|\E\left[r\left(O_t\right)-r(\tilde{O}_t)|\mathcal{F}_{t-\tau}\right]\right\|_1+\|\theta_{t-\tau}\|_\infty\left\|\E\left[\left(A\left(O_t\right)-A(\tilde{O}_t)\right)|\mathcal{F}_{t-\tau}\right]Q^{\pii_{t-\tau-1}}\right\|_1\\
&+ \|\theta_{t-\tau}\|_\infty\left\|\E\left[A\left(O_t\right)-A(\tilde{O}_t)|\mathcal{F}_{t-\tau}\right]\theta_{t-\tau}\right\|_1\\
\stackrel{(b)}{\leq}&\left\|\theta_{t-\tau}\right\|_\infty\left\|\E\left[r\left(O_t\right)-r(\tilde{O}_t)|\mathcal{F}_{t-\tau}\right]\right\|_1 + \|\theta_{t-\tau}\|_\infty\left\|\E\left[\left(A\left(O_t\right)-A(\tilde{O}_t)\right)|\mathcal{F}_{t-\tau}\right]\right\|_1\left\|Q^{\pii_{t-\tau-1}}\right\|_1\\
&+\left\|\theta_{t-\tau}\right\|_\infty.\left\|\E\left[A\left(O_t\right)-A(\tilde{O}_t)|\mathcal{F}_{t-\tau}\right]\right\|_1.\left\|\theta_{t-\tau}\right\|_1\\
\stackrel{(c)}{\leq}&2Q_{\max}\times 2|\mathcal{S}||\mathcal{A}|d_{TV}\left(O_t,\tilde{O}_t|\mathcal{F}_{t-\tau}\right) + 2Q_{\max}\times 2|\mathcal{S}||\mathcal{A}|d_{TV}\left(O_t,\tilde{O}_t|\mathcal{F}_{t-\tau}\right)\times Q_{\max}|\mathcal{S}||\mathcal{A}|\\
&+2Q_{\max}\times 2|\mathcal{S}||\mathcal{A}|d_{TV}\left(O_t,\tilde{O}_t|\mathcal{F}_{t-\tau}\right) \times   2Q_{\max}|\mathcal{S}||\mathcal{A}|\\
= & 4Q_{\max}|\mathcal{S}||\mathcal{A}|(1+3Q_{\max}|\mathcal{S}||\mathcal{A}|) d_{TV}\left(O_t,\tilde{O}_t|\mathcal{F}_{t-\tau}\right) 
\end{align*}
where $(a)$ is due to the H\"older's inequality, $(b)$ is due to definition of matrix induced norm, $(c)$ is due to Lemma \ref{lem:bounds}. Using Lemma \ref{lem:C.22}, we get the result.
\QEDopen

\textit{Proof of Lemma \ref{lem:D.7}:}
Consider the tuple $O'^t=(S_t',A_t',S_{t+1}',A_{t+1}')$, where $S_t'\sim \mu^{\pii_{t-\tau-1}}$, $A_t'\sim \pii_{t-\tau-1}(\cdot|S_t')$, $S_{t+1}'\sim \mathcal{P}(\cdot|S_t',A_t')$, and $A_{t+1}'\sim \pii_{t-\tau-1}(\cdot|S_{t+1}')$. We have
\begin{align*}
\E  \big[\Gamma(\pii_{t-\tau-1},\theta_{t-\tau},O'^t)\big|\mathcal{F}_{t-\tau}\big] 
=&\theta_{t-\tau}^\top\E\left[r(O_t')+A(O_t')Q^{\pii_{t-\tau-1}}|\mathcal{F}_{t-\tau}\right]\\
&+\theta_{t-\tau}^\top\E\left[A(O_t')|\mathcal{F}_{t-\tau}\right]\theta_{t-\tau}-\theta_{t-\tau}^\top\Bar{A}^{\pii_{t-\tau-1}}\theta_{t-\tau} = 0,
\end{align*}
where the last equality is due to the Bellman equation and the definition of $\Bar{A}^{\pi_{t-\tau-1}}$. As a result, we have:
\begin{align*}
\E  \big[\Gamma(\pii_{t-\tau-1}&,\theta_{t-\tau},\tilde{O}_t)\big|\mathcal{F}_{t-\tau}\big] \\
=&\E  \left[\Gamma(\pii_{t-\tau-1},\theta_{t-\tau},\tilde{O}_t)-\Gamma(\pii_{t-\tau-1},\theta_{t-\tau},O'^t)\big|\mathcal{F}_{t-\tau}\right] \\
\stackrel{(a)}{\leq} & \left\|\theta_{t-\tau}\right\|_{\infty}\left\|\E\left[r(\tilde{O}_t)-r\left(O_t'\right)|\mathcal{F}_{t-\tau}\right]\right\|_1+\left\|\theta_{t-\tau}\right\|_\infty\left\|\E\left[A(\tilde{O}_t)-A(O_t')\big|\mathcal{F}_{t-\tau}\right]\right\|_1\left\|Q^{\pii_{t-\tau-1}}\right\|_1\\
&+\left\|\theta_{t-\tau}\right\|_\infty.\left\|\E\left[A(\tilde{O}_t)-A(O_t')\big|\mathcal{F}_{t-\tau}\right]\right\|_1.\left\|\theta_{t-\tau}\right\|_1\\
\stackrel{(b)}{\leq} &2Q_{\max}\times 2|\mathcal{S}||\mathcal{A}|d_{TV}\left(\tilde{O}_t,O_t'|\mathcal{F}_{t-\tau}\right)+2Q_{\max}\times 2|\mathcal{S}||\mathcal{A}|d_{TV}(\tilde{O}_t,O_t'|\mathcal{F}_{t-\tau}) \times 3|\mathcal{S}||\mathcal{A}|Q_{\max}\\
\stackrel{(c)}{=}&C_b\sum_{s,a,s',a'}|P(\Tilde{S}_t=s|\mathcal{F}_{t-\tau})\pii_{t-\tau-1}(a|s)\mathcal{P}(s'|s,a)\pii_{t-\tau-1}(a'|s')\\
&-P(S_t'=s|\mathcal{F}_{t-\tau})\pii_{t-\tau-1}(a|s)\mathcal{P}(s'|s,a)\pii_{t-\tau-1}(a'|s')|\\
=&C_b\sum_{s,a,s',a'}\pii_{t-\tau-1}(a|s)\mathcal{P}(s'|s,a)\pii_{t-\tau-1}(a'|s')|P(\Tilde{S}_t=s|\mathcal{F}_{t-\tau})-P(S_t'=s|\mathcal{F}_{t-\tau})|\\
=&C_b \sum_{s}|P(\Tilde{S}_t=s|\mathcal{F}_{t-\tau})-P(S_t'=s|\mathcal{F}_{t-\tau})|\\
\stackrel{(d)}{\leq}& C_b m\rho^{\tau},
\end{align*}
where $(a)$ follows from H\"older's inequality and the definition of the matrix norm, $(b)$ follows from Lemma \ref{lem:bounds3}, in $(c)$ we defined $C_b=4Q_{\max}|\mathcal{S}||\mathcal{A}|(1+3|\mathcal{S}||\mathcal{A}|Q_{\max})$, and $(d)$ is due to the Lemma \ref{lem:mixing3}.
\QEDopen

\textit{Proof of Lemma \ref{lem:my_lemma2}:}
\begin{align*}
\sum_{i=t-\tau}^t\left\|\pii_i-\pii_{t-\tau-1}\right\|=&\sum_{i=t-\tau}^t\left\|\sum_{j=t-\tau}^i\pii_j-\pii_{j-1}\right\| \\
\stackrel{(a)}{\leq}& \sum_{i=t-\tau}^t\sum_{j=t-\tau}^i\left\|\pii_j-\pii_{j-1}\right\|\\
\stackrel{(b)}{\leq}& \sum_{i=t-\tau}^t\sum_{j=t-\tau}^i\left[\CLLL\frac{\epsilon_{j-2}}{j-1}+\CL \beta_{j-1}\right]
\leq (\tau+1)^2\left[\CLLL\frac{\epsilon_{t-\tau-2}}{t-\tau-1}+\CL\beta_{t-\tau-1}\right]
\end{align*}
where $(a)$ is by triangle inequality, and $(b)$ follows from Lemma \ref{lem:pi_lip3}.
\QEDopen

\textit{Proof of Lemma \ref{lem:bounds}:}
\begin{enumerate}
    \item The proof follows directly by Frobenius norm upper bound on the two norm of a matrix.
    \item Follows directly from assumption $\mathcal{R}(s,a)\leq 1 ~\forall s,a$.
    \item $\|\Bar{A}^\pi\|=\|\E_\pi A(O)\|\leq \E_\pi\|A(O)\|\leq \sqrt{2}$.
    \item $
    \left\|\E[r\left(O_1\right)-r(O_2)]\right\|_1 = \sum_{s,a}\left|\E\left(r(O_1)-r(O_2)\right)_{s,a}\right|$ $\leq  2|\mathcal{S}||\mathcal{A}|d_{TV}(O_1,O_2)$, where the inequality is due to $|r(O)_{s,a}|\leq 1$.
    \item $\left\|\E[A\left(O_1\right)-A(O_2)]\right\|_1\stackrel{(a)}{=}\max_{s',a'}\sum_{s,a}\left|\E\left(A(O_1)-A(O_2)\right)_{s,a,s',a'}\right|$ \newline $\stackrel{(b)}{\leq} \max_{s',a'}2|\mathcal{S}||\mathcal{A}|d_{TV}(O_1,O_2)=2|\mathcal{S}||\mathcal{A}|d_{TV}(O_1,O_2)$, \newline
    where $(a)$ is due to the definition of matrix norm, and $(b)$ is due to $|A(O)_{s,a,s',a'}|\leq 1$.
    \item $\|Q_{t+1}-Q_t\|\leq \sqrt{\sum_{s,a}\alpha_t^2(s,a)(2Q_{\max}+1)^2}$ $=\alpha_t(2Q_{\max}+1)$
    \item $\left\|\theta_t-\theta_{t-1}\right\|\leq \left\|Q_t-Q_{t-1}\right\|+\left\|Q^{\pii_{t-1}}-Q^{\pii_{t-2}}\right\|\leq  \Delta_Q\alpha_{t-1}+\CLL\left(\CLLL\frac{\epsilon_{t-3}}{t-2} +  \CL\beta_{t-2}\right)$\newline
    where the last inequality follows from the previous part, and Lemmas \ref{lem:cont4} and \ref{lem:pi_lip3}.\QEDopen  
\end{enumerate}

\textit{Proof of Lemma \ref{lem:C.22}:}
\begin{align*}
&d_{TV}(P(O_t\in\cdot\mid \mathcal{F}_{t-\tau})||P(\tilde{O}_t\in\cdot\mid\mathcal{F}_{t-\tau}))\\
=&\sum_{s,a,s',a'}\left|P(\overbrace{S_t=s,A_t=a}^{\mathcal{H}_t},S_{t+1}=s',A_{t+1}=a'|\mathcal{F}_{t-\tau})-P(\Tilde{S}_t=s,\Tilde{A}_t=a,\Tilde{S}_{t+1}=s',\Tilde{A}_{t+1}=a'|\mathcal{F}_{t-\tau})\right|\\
=&\sum_{s,a,s',a'}\big|\E[\pii_t(a'|s')|\mathcal{F}_{t-\tau},\mathcal{H}_t]\mathcal{P}(s'|s,a)P(S_t=s,A_t=a|\mathcal{F}_{t-\tau})\\
&- \pii_{t-\tau-1}(a'|s')\mathcal{P}(s'|s,a)P(\Tilde{S}_t=s,\Tilde{A}_t=a|\mathcal{F}_{t-\tau})\big|\\
\leq &\sum_{s,a,s',a'} \mathcal{P}(s'|s,a)P(S_t=s,A_t=a|\mathcal{F}_{t-\tau})\left|\E[\pii_t(a'|s')|\mathcal{F}_{t-\tau},\mathcal{H}_t]-\pii_{t-\tau-1}(a'|s')\right|\tag{$I_1$}\\
&+\sum_{s,a}
\big|P(S_t=s,A_t=a|\mathcal{F}_{t-\tau})-P(\Tilde{S}_t=s,\Tilde{A}_t=a|\mathcal{F}_{t-\tau})\big|\tag{$I_2$}.
\end{align*}
We bound $I_1$ and $I_2$ separately: 
\begin{align*}
I_1
&\leq \sum_{s,a,s',a'} \mathcal{P}(s'|s,a)P(S_t=s,A_t=a|\mathcal{F}_{t-\tau})\E[|\pii_t(a'|s')-\pii_{t-\tau-1}(a'|s')|\big|\mathcal{F}_{t-\tau},\mathcal{H}_t]\\
&\leq \sum_{s,a,s',a'} P(S_t=s,A_t=a|\mathcal{F}_{t-\tau})\E[|\pii_t(a'|s')-\pii_{t-\tau-1}(a'|s')|\big|\mathcal{F}_{t-\tau},\mathcal{H}_t]\\
&=\sum_{s',a'} \E[|\pii_t(a'|s')-\pii_{t-\tau-1}(a'|s')|\big|\mathcal{F}_{t-\tau}] \\
&\leq\sqrt{|\mathcal{A}||\mathcal{S}|}\E[\|\pii_t-\pii_{t-\tau-1}\|\big|\mathcal{F}_{t-\tau}],\\
I_2&=\sum_{s,a}\Bigg|\sum_{s'',a''}P(S_t=s,A_t=a,S_{t-1}=s'',A_{t-1}=a''|\mathcal{F}_{t-\tau})\\
&\quad\quad\quad-P(\Tilde{S}_t=s,\Tilde{A}_t=a,\Tilde{S}_{t-1}=s'',\Tilde{A}_{t-1}=a''|\mathcal{F}_{t-\tau})\Bigg|\\
&\leq \sum_{s,a,s'',a''}\bigg|P(S_t=s,A_t=a,S_{t-1}=s'',A_{t-1}=a''|\mathcal{F}_{t-\tau})\\
&~~~-P(\Tilde{S}_t=s,\Tilde{A}_t=a,\Tilde{S}_{t-1}=s'',\Tilde{A}_{t-1}=a''|\mathcal{F}_{t-\tau})\bigg|\\
&=d_{TV}(P(O_{t-1}\in\cdot|\mathcal{F}_{t-\tau})||P(\tilde{O}_{t-1}\in\cdot|\mathcal{F}_{t-\tau})).
\end{align*}
Combining the above bounds, we get:
\begin{align*}
d_{TV}(P(O_{t}\in\cdot|\mathcal{F}_{t-\tau})&||P(\tilde{O}_{t}\in\cdot|\mathcal{F}_{t-\tau})) \\
\leq & \sqrt{|\mathcal{A}||\mathcal{S}|}\E\left[\|\pii_t-\pii_{t-\tau-1}\|\bigg|\mathcal{F}_{t-\tau}\right]+d_{TV}(P(O_{t-1}\in\cdot|\mathcal{F}_{t-\tau})||P(\tilde{O}_{t-1}\in\cdot|\mathcal{F}_{t-\tau})).
\end{align*}
Following this induction, and noting that $P(S_{t-\tau}=s,A_{t-\tau}=a|\mathcal{F}_{t-\tau}) = P(\tilde{S}_{t-\tau}=s,\tilde{A}_{t-\tau}=a|\mathcal{F}_{t-\tau})$ (due to the definition of $\tilde{S}$ and $\tilde{A}$), we get the result.\QEDopen

\textit{Proof of Lemma \ref{lem:A.12}:} The proof follows directly from Lemma A.1 in \cite{wu2020finite}. \QEDopen

%% file: Thm_main_pf.tex
\section{Details of the Proof of Proposition \ref{thm:main2}} \label{sec:Thm_main_pf}
\subsection{Useful lemmas}
\textit{Proof of Lemma \ref{lem:logZ_bound3}:}
We have:
\begin{align}
\log Z_t(s)&=\log\sum_{a'}\pi_t(a|s)\exp(\beta_tQ_{t+1}(s,a)) \nonumber\\
&\geq \sum_{a}\pi_t(a|s)\beta_tQ_{t+1}(s,a),\label{eq:log_z_lower_bound}
\end{align}
where the inequality is due to the concavity of $\log(\cdot)$ function and Jensen's inequality. Furthermore, we have:
\begin{align}
V^{\pi_{t+1}}(\mu)-V^{\pi_{t}}(\mu) \stackrel{(a)}{=}&\frac{1}{1-\gamma}\sum_{s,a}d^{\pi_{t+1}}_\mu(s)\pi_{t+1}(a|s)\big[Q_{t+1}(s,a)+Q^{\pi_t}(s,a)-Q_{t+1}(s,a)-V^{\pi_t}(s)\big]\nonumber\\
\stackrel{(b)}{=}&\frac{1}{1-\gamma}\sum_{s,a}d^{\pi_{t+1}}_\mu(s)\pi_{t+1}(a|s)\bigg[\frac{1}{\beta_t}\log\frac{\pi_{t+1}(a|s)}{\pi_t(a|s)}\nonumber\\
&+\frac{1}{\beta_t}\log Z_t(s)+Q^{\pi_t}(s,a)-Q_{t+1}(s,a)-V^{\pi_t}(s)\bigg]\nonumber\\
\stackrel{(c)}{\geq} & \frac{1}{1-\gamma}\sum_{s,a}d^{\pi_{t+1}}_\mu(s)\pi_{t+1}(a|s)\bigg[\frac{1}{\beta_t}\log Z_t(s)+Q^{\pi_t}(s,a)-Q_{t+1}(s,a)-V^{\pi_t}(s)\bigg]\nonumber\\
=& \frac{1}{1-\gamma}\Bigg[\sum_{s,a}d^{\pi_{t+1}}_\mu(s)\pi_{t}(a|s)\left[\frac{1}{\beta_t}\log Z_t(s)-Q_{t+1}(s,a)\right]\nonumber\\
&+\sum_{s,a}d^{\pi_{t+1}}_\mu(s)(\pi_{t+1}(a|s)-\pi_t(a|s))\left[Q^{\pi_t}(s,a)-Q_{t+1}(s,a)\right]\Bigg]\nonumber\\
\stackrel{(d)}{\geq} & \sum_{s,a}\mu(s)\pi_{t}(a|s)\left[\frac{1}{\beta_t}\log Z_t(s)-Q_{t+1}(s,a)\right]\nonumber\\
&-\frac{2Q_{\max}\CL\sqrt{|\mathcal{A}|}}{1-\gamma}\beta_t \nonumber\\
=&\sum_{s}\mu(s)\left[\frac{1}{\beta_t}\log Z_t(s)-V^{\hat{\pi}_t}(s)\right]\nonumber\\
&+\sum_{s,a}\mu(s)\hat{\pi}_t(a|s)\left[Q^{\hat{\pi}_t}(s,a)-Q_{t+1}(s,a)\right]\nonumber\\
&+\sum_{s,a}\mu(s)(\hat{\pi}_t(a|s)-\pi_t(a|s))Q_{t+1}(s,a)\nonumber\\
&-\frac{2Q_{\max}\CL\sqrt{|\mathcal{A}|}}{1-\gamma}\beta_t,\nonumber
\end{align}
where $(a)$ is due to Performance Difference Lemma \cite{Kakade02approximatelyoptimal}, $(b)$ is by the update rule in Algorithm \ref{alg:1}, $(c)$ is by positivity of the KL-divergence \cite{cover2012elements}, and $(d)$ is by the definition of $d^{\pi_{t+1}}$ and \eqref{eq:log_z_lower_bound}. Taking $\mu=d^*$, we have:
\begin{align*}
    &\sum_{s}d^*(s)\left[\frac{1}{\beta_t}\log Z_t(s)-V^{\hat{\pi}_t}(s)\right]\nonumber
    \leq V^{\pi_{t+1}}(d^*)-V^{\pi_{t}}(d^*) + \|Q^{\pii_t}-Q_{t+1}\| + \frac{2\CL\sqrt{|\mathcal{A}|}}{(1-\gamma)^2}\beta_t + \frac{ \epsilon_{t-1}}{1-\gamma}
\end{align*}
which gets the result.\QEDopen

%% file: main.bbl
\begin{thebibliography}{70}

\bibitem{agarwal2019theory}
\begin{barticle}[author]
\bauthor{\bsnm{Agarwal},~\bfnm{Alekh}\binits{A.}},
  \bauthor{\bsnm{Kakade},~\bfnm{Sham~M}\binits{S.~M.}},
  \bauthor{\bsnm{Lee},~\bfnm{Jason~D}\binits{J.~D.}} \AND
  \bauthor{\bsnm{Mahajan},~\bfnm{Gaurav}\binits{G.}}
(\byear{2019}).
\btitle{On the theory of policy gradient methods: Optimality, approximation,
  and distribution shift}.
\bjournal{Preprint arXiv:1908.00261}.
\end{barticle}
\endbibitem

\bibitem{bahdanau2016actor}
\begin{barticle}[author]
\bauthor{\bsnm{Bahdanau},~\bfnm{Dzmitry}\binits{D.}},
  \bauthor{\bsnm{Brakel},~\bfnm{Philemon}\binits{P.}},
  \bauthor{\bsnm{Xu},~\bfnm{Kelvin}\binits{K.}},
  \bauthor{\bsnm{Goyal},~\bfnm{Anirudh}\binits{A.}},
  \bauthor{\bsnm{Lowe},~\bfnm{Ryan}\binits{R.}},
  \bauthor{\bsnm{Pineau},~\bfnm{Joelle}\binits{J.}},
  \bauthor{\bsnm{Courville},~\bfnm{Aaron}\binits{A.}} \AND
  \bauthor{\bsnm{Bengio},~\bfnm{Yoshua}\binits{Y.}}
(\byear{2016}).
\btitle{An actor-critic algorithm for sequence prediction}.
\bjournal{preprint arXiv:1607.07086}.
\end{barticle}
\endbibitem

\bibitem{bansal2019potential}
\begin{barticle}[author]
\bauthor{\bsnm{Bansal},~\bfnm{Nikhil}\binits{N.}} \AND
  \bauthor{\bsnm{Gupta},~\bfnm{Anupam}\binits{A.}}
(\byear{2019}).
\btitle{Potential-Function Proofs for Gradient Methods}.
\bjournal{Theory of Computing}
\bvolume{15}
\bpages{1--32}.
\end{barticle}
\endbibitem

\bibitem{beck2017first}
\begin{bbook}[author]
\bauthor{\bsnm{Beck},~\bfnm{Amir}\binits{A.}}
(\byear{2017}).
\btitle{First-order methods in optimization}.
\bpublisher{SIAM}.
\end{bbook}
\endbibitem

\bibitem{benveniste2012adaptive}
\begin{bbook}[author]
\bauthor{\bsnm{Benveniste},~\bfnm{Albert}\binits{A.}},
  \bauthor{\bsnm{M{\'e}tivier},~\bfnm{Michel}\binits{M.}} \AND
  \bauthor{\bsnm{Priouret},~\bfnm{Pierre}\binits{P.}}
(\byear{2012}).
\btitle{Adaptive algorithms and stochastic approximations}
\bvolume{22}.
\bpublisher{Springer Science \& Business Media}.
\end{bbook}
\endbibitem

\bibitem{Bhandari2018_FiniteTD}
\begin{binproceedings}[author]
\bauthor{\bsnm{Bhandari},~\bfnm{Jalaj}\binits{J.}},
  \bauthor{\bsnm{Russo},~\bfnm{Daniel}\binits{D.}} \AND
  \bauthor{\bsnm{Singal},~\bfnm{Raghav}\binits{R.}}
(\byear{2018}).
\btitle{A finite time analysis of temporal difference learning with linear
  function approximation}.
In \bbooktitle{Conference on learning theory}
\bpages{1691--1692}.
\bpublisher{PMLR}.
\end{binproceedings}
\endbibitem

\bibitem{bhatnagar2009natural}
\begin{barticle}[author]
\bauthor{\bsnm{Bhatnagar},~\bfnm{Shalabh}\binits{S.}},
  \bauthor{\bsnm{Sutton},~\bfnm{Richard~S}\binits{R.~S.}},
  \bauthor{\bsnm{Ghavamzadeh},~\bfnm{Mohammad}\binits{M.}} \AND
  \bauthor{\bsnm{Lee},~\bfnm{Mark}\binits{M.}}
(\byear{2009}).
\btitle{Natural actor--critic algorithms}.
\bjournal{Automatica}
\bvolume{45}
\bpages{2471--2482}.
\end{barticle}
\endbibitem

\bibitem{borkar2008}
\begin{bbook}[author]
\bauthor{\bsnm{Borkar},~\bfnm{V.~S.}\binits{V.~S.}}
(\byear{2008}).
\btitle{Stochastic Approximation: A Dynamical Systems Viewpoint}.
\bpublisher{Cambridge University Press}.
\end{bbook}
\endbibitem

\bibitem{borkar2009stochastic}
\begin{bbook}[author]
\bauthor{\bsnm{Borkar},~\bfnm{Vivek~S}\binits{V.~S.}}
(\byear{2009}).
\btitle{Stochastic approximation: a dynamical systems viewpoint}
\bvolume{48}.
\bpublisher{Springer}.
\end{bbook}
\endbibitem

\bibitem{borkar2018concentration}
\begin{binproceedings}[author]
\bauthor{\bsnm{Borkar},~\bfnm{Vivek~S}\binits{V.~S.}} \AND
  \bauthor{\bsnm{Pattathil},~\bfnm{Sarath}\binits{S.}}
(\byear{2018}).
\btitle{Concentration bounds for two time scale stochastic approximation}.
In \bbooktitle{2018 56th Annual Allerton Conference on Communication, Control,
  and Computing (Allerton)}
\bpages{504--511}.
\bpublisher{IEEE}.
\end{binproceedings}
\endbibitem

\bibitem{bouneffouf2012contextual}
\begin{binproceedings}[author]
\bauthor{\bsnm{Bouneffouf},~\bfnm{Djallel}\binits{D.}},
  \bauthor{\bsnm{Bouzeghoub},~\bfnm{Amel}\binits{A.}} \AND
  \bauthor{\bsnm{Gan{\c{c}}arski},~\bfnm{Alda~Lopes}\binits{A.~L.}}
(\byear{2012}).
\btitle{A contextual-bandit algorithm for mobile context-aware recommender
  system}.
In \bbooktitle{International conference on neural information processing}
\bpages{324--331}.
\bpublisher{Springer}.
\end{binproceedings}
\endbibitem

\bibitem{cayci2021linear}
\begin{barticle}[author]
\bauthor{\bsnm{Cayci},~\bfnm{Semih}\binits{S.}},
  \bauthor{\bsnm{He},~\bfnm{Niao}\binits{N.}} \AND
  \bauthor{\bsnm{Srikant},~\bfnm{R}\binits{R.}}
(\byear{2021}).
\btitle{Linear Convergence of Entropy-Regularized Natural Policy Gradient with
  Linear Function Approximation}.
\bjournal{arXiv preprint arXiv:2106.04096}.
\end{barticle}
\endbibitem

\bibitem{cen2020fast}
\begin{barticle}[author]
\bauthor{\bsnm{Cen},~\bfnm{Shicong}\binits{S.}},
  \bauthor{\bsnm{Cheng},~\bfnm{Chen}\binits{C.}},
  \bauthor{\bsnm{Chen},~\bfnm{Yuxin}\binits{Y.}},
  \bauthor{\bsnm{Wei},~\bfnm{Yuting}\binits{Y.}} \AND
  \bauthor{\bsnm{Chi},~\bfnm{Yuejie}\binits{Y.}}
(\byear{2020}).
\btitle{Fast global convergence of natural policy gradient methods with entropy
  regularization}.
\bjournal{arXiv preprint arXiv:2007.06558}.
\end{barticle}
\endbibitem

\bibitem{chen2021finite}
\begin{barticle}[author]
\bauthor{\bsnm{Chen},~\bfnm{Zaiwei}\binits{Z.}},
  \bauthor{\bsnm{Khodadadian},~\bfnm{Sajad}\binits{S.}} \AND
  \bauthor{\bsnm{Maguluri},~\bfnm{Siva~Theja}\binits{S.~T.}}
(\byear{2021}).
\btitle{Finite-Sample Analysis of Off-Policy Natural Actor-Critic with Linear
  Function Approximation}.
\bjournal{arXiv preprint arXiv:2105.12540}.
\end{barticle}
\endbibitem

\bibitem{chen2020finite}
\begin{barticle}[author]
\bauthor{\bsnm{Chen},~\bfnm{Zaiwei}\binits{Z.}},
  \bauthor{\bsnm{Maguluri},~\bfnm{Siva~Theja}\binits{S.~T.}},
  \bauthor{\bsnm{Shakkottai},~\bfnm{Sanjay}\binits{S.}} \AND
  \bauthor{\bsnm{Shanmugam},~\bfnm{Karthikeyan}\binits{K.}}
(\byear{2020}).
\btitle{Finite-Sample Analysis of Contractive Stochastic Approximation Using
  Smooth Convex Envelopes}.
\bjournal{Advances in Neural Information Processing Systems}
\bvolume{33}.
\end{barticle}
\endbibitem

\bibitem{cover2012elements}
\begin{bbook}[author]
\bauthor{\bsnm{Cover},~\bfnm{Thomas~M}\binits{T.~M.}} \AND
  \bauthor{\bsnm{Thomas},~\bfnm{Joy~A}\binits{J.~A.}}
(\byear{2012}).
\btitle{Elements of information theory}.
\bpublisher{John Wiley \& Sons}.
\end{bbook}
\endbibitem

\bibitem{davis1961markov}
\begin{barticle}[author]
\bauthor{\bsnm{Davis},~\bfnm{AS}\binits{A.}}
(\byear{1961}).
\btitle{Markov chains as random input automata}.
\bjournal{The American Mathematical Monthly}
\bvolume{68}
\bpages{264--267}.
\end{barticle}
\endbibitem

\bibitem{doan2019finite}
\begin{barticle}[author]
\bauthor{\bsnm{Doan},~\bfnm{Thinh~T}\binits{T.~T.}}
(\byear{2019}).
\btitle{Finite-Time Analysis and Restarting Scheme for Linear Two-Time-Scale
  Stochastic Approximation}.
\bjournal{preprint arXiv:1912.10583}.
\end{barticle}
\endbibitem

\bibitem{doan2021finite}
\begin{barticle}[author]
\bauthor{\bsnm{Doan},~\bfnm{Thinh~T}\binits{T.~T.}}
(\byear{2021}).
\btitle{Finite-Time Convergence Rates of Nonlinear Two-Time-Scale Stochastic
  Approximation under Markovian Noise}.
\bjournal{arXiv preprint arXiv:2104.01627}.
\end{barticle}
\endbibitem

\bibitem{espeholt2018impala}
\begin{barticle}[author]
\bauthor{\bsnm{Espeholt},~\bfnm{Lasse}\binits{L.}},
  \bauthor{\bsnm{Soyer},~\bfnm{Hubert}\binits{H.}},
  \bauthor{\bsnm{Munos},~\bfnm{Remi}\binits{R.}},
  \bauthor{\bsnm{Simonyan},~\bfnm{Karen}\binits{K.}},
  \bauthor{\bsnm{Mnih},~\bfnm{Volodymir}\binits{V.}},
  \bauthor{\bsnm{Ward},~\bfnm{Tom}\binits{T.}},
  \bauthor{\bsnm{Doron},~\bfnm{Yotam}\binits{Y.}},
  \bauthor{\bsnm{Firoiu},~\bfnm{Vlad}\binits{V.}},
  \bauthor{\bsnm{Harley},~\bfnm{Tim}\binits{T.}},
  \bauthor{\bsnm{Dunning},~\bfnm{Iain}\binits{I.}} \betal{et~al.}
(\byear{2018}).
\btitle{Impala: Scalable distributed deep-RL with importance weighted
  actor-learner architectures}.
\bjournal{arXiv preprint arXiv:1802.01561}.
\end{barticle}
\endbibitem

\bibitem{fu2020single}
\begin{barticle}[author]
\bauthor{\bsnm{Fu},~\bfnm{Zuyue}\binits{Z.}},
  \bauthor{\bsnm{Yang},~\bfnm{Zhuoran}\binits{Z.}} \AND
  \bauthor{\bsnm{Wang},~\bfnm{Zhaoran}\binits{Z.}}
(\byear{2020}).
\btitle{Single-Timescale Actor-Critic Provably Finds Globally Optimal Policy}.
\bjournal{arXiv preprint arXiv:2008.00483}.
\end{barticle}
\endbibitem

\bibitem{Gajjar_2003}
\begin{barticle}[author]
\bauthor{\bsnm{Gajjar},~\bfnm{GR}\binits{G.}},
  \bauthor{\bsnm{Khaparde},~\bfnm{SA}\binits{S.}},
  \bauthor{\bsnm{Nagaraju},~\bfnm{P}\binits{P.}} \AND
  \bauthor{\bsnm{Soman},~\bfnm{SA}\binits{S.}}
(\byear{2003}).
\btitle{Application of actor-critic learning algorithm for optimal bidding
  problem of a Genco}.
\bjournal{IEEE Transactions on Power Systems}
\bvolume{18}
\bpages{11--18}.
\end{barticle}
\endbibitem

\bibitem{geist2019theory}
\begin{barticle}[author]
\bauthor{\bsnm{Geist},~\bfnm{Matthieu}\binits{M.}},
  \bauthor{\bsnm{Scherrer},~\bfnm{Bruno}\binits{B.}} \AND
  \bauthor{\bsnm{Pietquin},~\bfnm{Olivier}\binits{O.}}
(\byear{2019}).
\btitle{A theory of regularized markov decision processes}.
\bjournal{preprint arXiv:1901.11275}.
\end{barticle}
\endbibitem

\bibitem{gunasekar2020mirrorless}
\begin{barticle}[author]
\bauthor{\bsnm{Gunasekar},~\bfnm{Suriya}\binits{S.}},
  \bauthor{\bsnm{Woodworth},~\bfnm{Blake}\binits{B.}} \AND
  \bauthor{\bsnm{Srebro},~\bfnm{Nathan}\binits{N.}}
(\byear{2020}).
\btitle{Mirrorless Mirror Descent: A More Natural Discretization of Riemannian
  Gradient Flow}.
\bjournal{arXiv preprint arXiv:2004.01025}.
\end{barticle}
\endbibitem

\bibitem{Haarnoja_2019}
\begin{barticle}[author]
\bauthor{\bsnm{Haarnoja},~\bfnm{Tuomas}\binits{T.}},
  \bauthor{\bsnm{Zhou},~\bfnm{Aurick}\binits{A.}},
  \bauthor{\bsnm{Hartikainen},~\bfnm{Kristian}\binits{K.}},
  \bauthor{\bsnm{Tucker},~\bfnm{George}\binits{G.}},
  \bauthor{\bsnm{Ha},~\bfnm{Sehoon}\binits{S.}},
  \bauthor{\bsnm{Tan},~\bfnm{Jie}\binits{J.}},
  \bauthor{\bsnm{Kumar},~\bfnm{Vikash}\binits{V.}},
  \bauthor{\bsnm{Zhu},~\bfnm{Henry}\binits{H.}},
  \bauthor{\bsnm{Gupta},~\bfnm{Abhishek}\binits{A.}},
  \bauthor{\bsnm{Abbeel},~\bfnm{Pieter}\binits{P.}} \betal{et~al.}
(\byear{2018}).
\btitle{Soft actor-critic algorithms and applications}.
\bjournal{preprint arXiv:1812.05905}.
\end{barticle}
\endbibitem

\bibitem{hajek2015random}
\begin{bbook}[author]
\bauthor{\bsnm{Hajek},~\bfnm{Bruce}\binits{B.}}
(\byear{2015}).
\btitle{Random processes for engineers}.
\bpublisher{Cambridge university press}.
\end{bbook}
\endbibitem

\bibitem{Kakade02approximatelyoptimal}
\begin{binproceedings}[author]
\bauthor{\bsnm{Kakade},~\bfnm{Sham}\binits{S.}} \AND
  \bauthor{\bsnm{Langford},~\bfnm{John}\binits{J.}}
(\byear{2002}).
\btitle{Approximately optimal approximate reinforcement learning}.
In \bbooktitle{ICML}
\bvolume{2}
\bpages{267--274}.
\end{binproceedings}
\endbibitem

\bibitem{kakade2002natural}
\begin{binproceedings}[author]
\bauthor{\bsnm{Kakade},~\bfnm{Sham~M}\binits{S.~M.}}
(\byear{2002}).
\btitle{A natural policy gradient}.
In \bbooktitle{Advances in neural information processing systems}
\bpages{1531--1538}.
\end{binproceedings}
\endbibitem

\bibitem{kaledin2020finite}
\begin{binproceedings}[author]
\bauthor{\bsnm{Kaledin},~\bfnm{Maxim}\binits{M.}},
  \bauthor{\bsnm{Moulines},~\bfnm{Eric}\binits{E.}},
  \bauthor{\bsnm{Naumov},~\bfnm{Alexey}\binits{A.}},
  \bauthor{\bsnm{Tadic},~\bfnm{Vladislav}\binits{V.}} \AND
  \bauthor{\bsnm{Wai},~\bfnm{Hoi-To}\binits{H.-T.}}
(\byear{2020}).
\btitle{Finite time analysis of linear two-timescale stochastic approximation
  with Markovian noise}.
In \bbooktitle{Conference on Learning Theory}
\bpages{2144--2203}.
\bpublisher{PMLR}.
\end{binproceedings}
\endbibitem

\bibitem{Kaledin_two_time_scale_2020}
\begin{barticle}[author]
\bauthor{\bsnm{Kaledin},~\bfnm{Maxim}\binits{M.}},
  \bauthor{\bsnm{Moulines},~\bfnm{Eric}\binits{E.}},
  \bauthor{\bsnm{Naumov},~\bfnm{Alexey}\binits{A.}},
  \bauthor{\bsnm{Tadic},~\bfnm{Vladislav}\binits{V.}} \AND
  \bauthor{\bsnm{Wai},~\bfnm{Hoi-To}\binits{H.-T.}}
(\byear{2020}).
\btitle{Finite Time Analysis of Linear Two-timescale Stochastic Approximation
  with Markovian Noise}.
\bjournal{preprint arXiv:2002.01268}.
\end{barticle}
\endbibitem

\bibitem{khodadadian2021finite}
\begin{barticle}[author]
\bauthor{\bsnm{Khodadadian},~\bfnm{Sajad}\binits{S.}},
  \bauthor{\bsnm{Chen},~\bfnm{Zaiwei}\binits{Z.}} \AND
  \bauthor{\bsnm{Maguluri},~\bfnm{Siva~Theja}\binits{S.~T.}}
(\byear{2021}).
\btitle{Finite-Sample Analysis of Off-Policy Natural Actor-Critic Algorithm}.
\bjournal{arXiv preprint arXiv:2102.09318}.
\end{barticle}
\endbibitem

\bibitem{khodadadian2021linear}
\begin{barticle}[author]
\bauthor{\bsnm{Khodadadian},~\bfnm{Sajad}\binits{S.}},
  \bauthor{\bsnm{Jhunjhunwala},~\bfnm{Prakirt~Raj}\binits{P.~R.}},
  \bauthor{\bsnm{Varma},~\bfnm{Sushil~Mahavir}\binits{S.~M.}} \AND
  \bauthor{\bsnm{Maguluri},~\bfnm{Siva~Theja}\binits{S.~T.}}
(\byear{2021}).
\btitle{On the linear convergence of natural policy gradient algorithm}.
\bjournal{arXiv preprint arXiv:2105.01424}.
\end{barticle}
\endbibitem

\bibitem{konda2000actor}
\begin{binproceedings}[author]
\bauthor{\bsnm{Konda},~\bfnm{Vijay~R}\binits{V.~R.}} \AND
  \bauthor{\bsnm{Tsitsiklis},~\bfnm{John~N}\binits{J.~N.}}
(\byear{2000}).
\btitle{Actor-critic algorithms}.
In \bbooktitle{Advances in neural information processing systems}
\bpages{1008--1014}.
\end{binproceedings}
\endbibitem

\bibitem{kuleshov2014algorithms}
\begin{barticle}[author]
\bauthor{\bsnm{Kuleshov},~\bfnm{Volodymyr}\binits{V.}} \AND
  \bauthor{\bsnm{Precup},~\bfnm{Doina}\binits{D.}}
(\byear{2014}).
\btitle{Algorithms for multi-armed bandit problems}.
\bjournal{arXiv preprint arXiv:1402.6028}.
\end{barticle}
\endbibitem

\bibitem{kumar2019sample}
\begin{barticle}[author]
\bauthor{\bsnm{Kumar},~\bfnm{Harshat}\binits{H.}},
  \bauthor{\bsnm{Koppel},~\bfnm{Alec}\binits{A.}} \AND
  \bauthor{\bsnm{Ribeiro},~\bfnm{Alejandro}\binits{A.}}
(\byear{2019}).
\btitle{On the Sample Complexity of Actor-Critic Method for Reinforcement
  Learning with Function Approximation}.
\bjournal{preprint arXiv:1910.08412}.
\end{barticle}
\endbibitem

\bibitem{lan2021policy}
\begin{barticle}[author]
\bauthor{\bsnm{Lan},~\bfnm{Guanghui}\binits{G.}}
(\byear{2021}).
\btitle{Policy mirror descent for reinforcement learning: Linear convergence,
  new sampling complexity, and generalized problem classes}.
\bjournal{arXiv preprint arXiv:2102.00135}.
\end{barticle}
\endbibitem

\bibitem{levin2017markov}
\begin{bbook}[author]
\bauthor{\bsnm{Levin},~\bfnm{David~A}\binits{D.~A.}} \AND
  \bauthor{\bsnm{Peres},~\bfnm{Yuval}\binits{Y.}}
(\byear{2017}).
\btitle{Markov chains and mixing times}
\bvolume{107}.
\bpublisher{American Mathematical Soc.}
\end{bbook}
\endbibitem

\bibitem{liu2019neural}
\begin{barticle}[author]
\bauthor{\bsnm{Liu},~\bfnm{Boyi}\binits{B.}},
  \bauthor{\bsnm{Cai},~\bfnm{Qi}\binits{Q.}},
  \bauthor{\bsnm{Yang},~\bfnm{Zhuoran}\binits{Z.}} \AND
  \bauthor{\bsnm{Wang},~\bfnm{Zhaoran}\binits{Z.}}
(\byear{2019}).
\btitle{Neural proximal/trust region policy optimization attains globally
  optimal policy}.
\bjournal{Advances in Neural Information Processing Systems}
\bvolume{32}.
\end{barticle}
\endbibitem

\bibitem{liu2015finite}
\begin{binproceedings}[author]
\bauthor{\bsnm{Liu},~\bfnm{Bo}\binits{B.}},
  \bauthor{\bsnm{Liu},~\bfnm{Ji}\binits{J.}},
  \bauthor{\bsnm{Ghavamzadeh},~\bfnm{Mohammad}\binits{M.}},
  \bauthor{\bsnm{Mahadevan},~\bfnm{Sridhar}\binits{S.}} \AND
  \bauthor{\bsnm{Petrik},~\bfnm{Marek}\binits{M.}}
(\byear{2015}).
\btitle{Finite-Sample Analysis of Proximal Gradient TD Algorithms.}
In \bbooktitle{UAI}
\bpages{504--513}.
\bpublisher{Citeseer}.
\end{binproceedings}
\endbibitem

\bibitem{liu2020improved}
\begin{barticle}[author]
\bauthor{\bsnm{Liu},~\bfnm{Yanli}\binits{Y.}},
  \bauthor{\bsnm{Zhang},~\bfnm{Kaiqing}\binits{K.}},
  \bauthor{\bsnm{Basar},~\bfnm{Tamer}\binits{T.}} \AND
  \bauthor{\bsnm{Yin},~\bfnm{Wotao}\binits{W.}}
(\byear{2020}).
\btitle{An improved analysis of (variance-reduced) policy gradient and natural
  policy gradient methods}.
\bjournal{Advances in Neural Information Processing Systems}
\bvolume{33}.
\end{barticle}
\endbibitem

\bibitem{mei2020global}
\begin{barticle}[author]
\bauthor{\bsnm{Mei},~\bfnm{Jincheng}\binits{J.}},
  \bauthor{\bsnm{Xiao},~\bfnm{Chenjun}\binits{C.}},
  \bauthor{\bsnm{Szepesvari},~\bfnm{Csaba}\binits{C.}} \AND
  \bauthor{\bsnm{Schuurmans},~\bfnm{Dale}\binits{D.}}
(\byear{2020}).
\btitle{On the Global Convergence Rates of Softmax Policy Gradient Methods}.
\bjournal{preprint arXiv:2005.06392}.
\end{barticle}
\endbibitem

\bibitem{montague1999reinforcement}
\begin{barticle}[author]
\bauthor{\bsnm{Montague},~\bfnm{P~Read}\binits{P.~R.}}
(\byear{1999}).
\btitle{Reinforcement Learning: An Introduction, by Sutton, RS and Barto, AG}.
\bjournal{Trends in cognitive sciences}
\bvolume{3}
\bpages{360}.
\end{barticle}
\endbibitem

\bibitem{NIPS2009_3767}
\begin{binproceedings}[author]
\bauthor{\bsnm{Morimura},~\bfnm{Tetsuro}\binits{T.}},
  \bauthor{\bsnm{Uchibe},~\bfnm{Eiji}\binits{E.}},
  \bauthor{\bsnm{Yoshimoto},~\bfnm{Junichiro}\binits{J.}} \AND
  \bauthor{\bsnm{Doya},~\bfnm{Kenji}\binits{K.}}
(\byear{2009}).
\btitle{A generalized natural actor-critic algorithm}.
In \bbooktitle{Advances in neural information processing systems}
\bpages{1312--1320}.
\end{binproceedings}
\endbibitem

\bibitem{mou2020linear}
\begin{binproceedings}[author]
\bauthor{\bsnm{Mou},~\bfnm{Wenlong}\binits{W.}},
  \bauthor{\bsnm{Li},~\bfnm{Chris~Junchi}\binits{C.~J.}},
  \bauthor{\bsnm{Wainwright},~\bfnm{Martin~J}\binits{M.~J.}},
  \bauthor{\bsnm{Bartlett},~\bfnm{Peter~L}\binits{P.~L.}} \AND
  \bauthor{\bsnm{Jordan},~\bfnm{Michael~I}\binits{M.~I.}}
(\byear{2020}).
\btitle{On linear stochastic approximation: Fine-grained Polyak-Ruppert and
  non-asymptotic concentration}.
In \bbooktitle{Conference on Learning Theory}
\bpages{2947--2997}.
\bpublisher{PMLR}.
\end{binproceedings}
\endbibitem

\bibitem{peters2008natural}
\begin{barticle}[author]
\bauthor{\bsnm{Peters},~\bfnm{Jan}\binits{J.}} \AND
  \bauthor{\bsnm{Schaal},~\bfnm{Stefan}\binits{S.}}
(\byear{2008}).
\btitle{Natural actor-critic}.
\bjournal{Neurocomputing}
\bvolume{71}
\bpages{1180--1190}.
\end{barticle}
\endbibitem

\bibitem{puterman1990markov}
\begin{barticle}[author]
\bauthor{\bsnm{Puterman},~\bfnm{Martin~L}\binits{M.~L.}}
(\byear{1990}).
\btitle{Markov decision processes}.
\bjournal{Handbooks in operations research and management science}
\bvolume{2}
\bpages{331--434}.
\end{barticle}
\endbibitem

\bibitem{qiu2019finite}
\begin{binproceedings}[author]
\bauthor{\bsnm{Qiu},~\bfnm{Shuang}\binits{S.}},
  \bauthor{\bsnm{Yang},~\bfnm{Zhuoran}\binits{Z.}},
  \bauthor{\bsnm{Ye},~\bfnm{Jieping}\binits{J.}} \AND
  \bauthor{\bsnm{Wang},~\bfnm{Zhaoran}\binits{Z.}}
(\byear{2019}).
\btitle{On the finite-time convergence of actor-critic algorithm}.
In \bbooktitle{Optimization Foundations for Reinforcement Learning Workshop at
  Advances in Neural Information Processing Systems (NeurIPS)}.
\end{binproceedings}
\endbibitem

\bibitem{raskutti2015information}
\begin{barticle}[author]
\bauthor{\bsnm{Raskutti},~\bfnm{Garvesh}\binits{G.}} \AND
  \bauthor{\bsnm{Mukherjee},~\bfnm{Sayan}\binits{S.}}
(\byear{2015}).
\btitle{The information geometry of mirror descent}.
\bjournal{IEEE Transactions on Information Theory}
\bvolume{61}
\bpages{1451--1457}.
\end{barticle}
\endbibitem

\bibitem{rissanen1996fisher}
\begin{barticle}[author]
\bauthor{\bsnm{Rissanen},~\bfnm{Jorma~J}\binits{J.~J.}}
(\byear{1996}).
\btitle{Fisher information and stochastic complexity}.
\bjournal{IEEE transactions on information theory}
\bvolume{42}
\bpages{40--47}.
\end{barticle}
\endbibitem

\bibitem{robbins1951stochastic}
\begin{barticle}[author]
\bauthor{\bsnm{Robbins},~\bfnm{Herbert}\binits{H.}} \AND
  \bauthor{\bsnm{Monro},~\bfnm{Sutton}\binits{S.}}
(\byear{1951}).
\btitle{A stochastic approximation method}.
\bjournal{The annals of mathematical statistics}
\bpages{400--407}.
\end{barticle}
\endbibitem

\bibitem{schulman2015trust}
\begin{binproceedings}[author]
\bauthor{\bsnm{Schulman},~\bfnm{John}\binits{J.}},
  \bauthor{\bsnm{Levine},~\bfnm{Sergey}\binits{S.}},
  \bauthor{\bsnm{Abbeel},~\bfnm{Pieter}\binits{P.}},
  \bauthor{\bsnm{Jordan},~\bfnm{Michael}\binits{M.}} \AND
  \bauthor{\bsnm{Moritz},~\bfnm{Philipp}\binits{P.}}
(\byear{2015}).
\btitle{Trust region policy optimization}.
In \bbooktitle{International conference on machine learning}
\bpages{1889--1897}.
\end{binproceedings}
\endbibitem

\bibitem{schulman2017proximal}
\begin{barticle}[author]
\bauthor{\bsnm{Schulman},~\bfnm{John}\binits{J.}},
  \bauthor{\bsnm{Wolski},~\bfnm{Filip}\binits{F.}},
  \bauthor{\bsnm{Dhariwal},~\bfnm{Prafulla}\binits{P.}},
  \bauthor{\bsnm{Radford},~\bfnm{Alec}\binits{A.}} \AND
  \bauthor{\bsnm{Klimov},~\bfnm{Oleg}\binits{O.}}
(\byear{2017}).
\btitle{Proximal policy optimization algorithms}.
\bjournal{preprint arXiv:1707.06347}.
\end{barticle}
\endbibitem

\bibitem{shani2019adaptive}
\begin{barticle}[author]
\bauthor{\bsnm{Shani},~\bfnm{Lior}\binits{L.}},
  \bauthor{\bsnm{Efroni},~\bfnm{Yonathan}\binits{Y.}} \AND
  \bauthor{\bsnm{Mannor},~\bfnm{Shie}\binits{S.}}
(\byear{2019}).
\btitle{Adaptive trust region policy optimization: Global convergence and
  faster rates for regularized {MDP}s}.
\bjournal{arXiv preprint arXiv:1909.02769}.
\end{barticle}
\endbibitem

\bibitem{shani2020adaptive}
\begin{binproceedings}[author]
\bauthor{\bsnm{Shani},~\bfnm{Lior}\binits{L.}},
  \bauthor{\bsnm{Efroni},~\bfnm{Yonathan}\binits{Y.}} \AND
  \bauthor{\bsnm{Mannor},~\bfnm{Shie}\binits{S.}}
(\byear{2020}).
\btitle{{Adaptive Trust Region Policy Optimization: Global Convergence and
  Faster Rates for Regularized MDPs}}.
In \bbooktitle{Proceedings of the AAAI Conference on Artificial Intelligence}
\bvolume{34}
\bpages{5668--5675}.
\end{binproceedings}
\endbibitem

\bibitem{sutton2000policy}
\begin{binproceedings}[author]
\bauthor{\bsnm{Sutton},~\bfnm{Richard~S}\binits{R.~S.}},
  \bauthor{\bsnm{McAllester},~\bfnm{David~A}\binits{D.~A.}},
  \bauthor{\bsnm{Singh},~\bfnm{Satinder~P}\binits{S.~P.}} \AND
  \bauthor{\bsnm{Mansour},~\bfnm{Yishay}\binits{Y.}}
(\byear{2000}).
\btitle{Policy gradient methods for reinforcement learning with function
  approximation}.
In \bbooktitle{Advances in neural information processing systems}
\bpages{1057--1063}.
\end{binproceedings}
\endbibitem

\bibitem{NIPS2013_5184}
\begin{bincollection}[author]
\bauthor{\bsnm{Thomas},~\bfnm{Philip~S.}\binits{P.~S.}},
  \bauthor{\bsnm{Dabney},~\bfnm{William~C}\binits{W.~C.}},
  \bauthor{\bsnm{Giguere},~\bfnm{Stephen}\binits{S.}} \AND
  \bauthor{\bsnm{Mahadevan},~\bfnm{Sridhar}\binits{S.}}
(\byear{2013}).
\btitle{Projected Natural Actor-Critic}.
In \bbooktitle{Advances in Neural Information Processing Systems 26}
(\beditor{\bfnm{C.~J.~C.}\binits{C.~J.~C.}~\bsnm{Burges}},
  \beditor{\bfnm{L.}\binits{L.}~\bsnm{Bottou}},
  \beditor{\bfnm{M.}\binits{M.}~\bsnm{Welling}},
  \beditor{\bfnm{Z.}\binits{Z.}~\bsnm{Ghahramani}} \AND
  \beditor{\bfnm{K.~Q.}\binits{K.~Q.}~\bsnm{Weinberger}}, eds.)
\bpages{2337--2345}.
\bpublisher{Curran Associates, Inc.}
\end{bincollection}
\endbibitem

\bibitem{tokic2010adaptive}
\begin{binproceedings}[author]
\bauthor{\bsnm{Tokic},~\bfnm{Michel}\binits{M.}}
(\byear{2010}).
\btitle{Adaptive $\varepsilon$-greedy exploration in reinforcement learning
  based on value differences}.
In \bbooktitle{Annual Conference on Artificial Intelligence}
\bpages{203--210}.
\bpublisher{Springer}.
\end{binproceedings}
\endbibitem

\bibitem{wang2019neural}
\begin{barticle}[author]
\bauthor{\bsnm{Wang},~\bfnm{Lingxiao}\binits{L.}},
  \bauthor{\bsnm{Cai},~\bfnm{Qi}\binits{Q.}},
  \bauthor{\bsnm{Yang},~\bfnm{Zhuoran}\binits{Z.}} \AND
  \bauthor{\bsnm{Wang},~\bfnm{Zhaoran}\binits{Z.}}
(\byear{2019}).
\btitle{Neural policy gradient methods: Global optimality and rates of
  convergence}.
\bjournal{preprint arXiv:1909.01150}.
\end{barticle}
\endbibitem

\bibitem{wang2016sample}
\begin{barticle}[author]
\bauthor{\bsnm{Wang},~\bfnm{Ziyu}\binits{Z.}},
  \bauthor{\bsnm{Bapst},~\bfnm{Victor}\binits{V.}},
  \bauthor{\bsnm{Heess},~\bfnm{Nicolas}\binits{N.}},
  \bauthor{\bsnm{Mnih},~\bfnm{Volodymyr}\binits{V.}},
  \bauthor{\bsnm{Munos},~\bfnm{Remi}\binits{R.}},
  \bauthor{\bsnm{Kavukcuoglu},~\bfnm{Koray}\binits{K.}} \AND
  \bauthor{\bparticle{de} \bsnm{Freitas},~\bfnm{Nando}\binits{N.}}
(\byear{2016}).
\btitle{Sample efficient actor-critic with experience replay}.
\bjournal{preprint arXiv:1611.01224}.
\end{barticle}
\endbibitem

\bibitem{watkins1992q}
\begin{barticle}[author]
\bauthor{\bsnm{Watkins},~\bfnm{Christopher~JCH}\binits{C.~J.}} \AND
  \bauthor{\bsnm{Dayan},~\bfnm{Peter}\binits{P.}}
(\byear{1992}).
\btitle{Q-learning}.
\bjournal{Machine learning}
\bvolume{8}
\bpages{279--292}.
\end{barticle}
\endbibitem

\bibitem{williams1990mathematical}
\begin{binproceedings}[author]
\bauthor{\bsnm{Williams},~\bfnm{Ronald~J}\binits{R.~J.}} \AND
  \bauthor{\bsnm{Baird},~\bfnm{Leemon~C}\binits{L.~C.}}
(\byear{1990}).
\btitle{A mathematical analysis of actor-critic architectures for learning
  optimal controls through incremental dynamic programming}.
In \bbooktitle{Proceedings of the Sixth Yale Workshop on Adaptive and Learning
  Systems}
\bpages{96--101}.
\bpublisher{Citeseer}.
\end{binproceedings}
\endbibitem

\bibitem{wu2020finite}
\begin{barticle}[author]
\bauthor{\bsnm{Wu},~\bfnm{Yue}\binits{Y.}},
  \bauthor{\bsnm{Zhang},~\bfnm{Weitong}\binits{W.}},
  \bauthor{\bsnm{Xu},~\bfnm{Pan}\binits{P.}} \AND
  \bauthor{\bsnm{Gu},~\bfnm{Quanquan}\binits{Q.}}
(\byear{2020}).
\btitle{A Finite Time Analysis of Two Time-Scale Actor Critic Methods}.
\bjournal{preprint arXiv:2005.01350}.
\end{barticle}
\endbibitem

\bibitem{wunder2010classes}
\begin{binproceedings}[author]
\bauthor{\bsnm{Wunder},~\bfnm{Michael}\binits{M.}},
  \bauthor{\bsnm{Littman},~\bfnm{Michael~L}\binits{M.~L.}} \AND
  \bauthor{\bsnm{Babes},~\bfnm{Monica}\binits{M.}}
(\byear{2010}).
\btitle{Classes of multiagent q-learning dynamics with epsilon-greedy
  exploration}.
In \bbooktitle{Proceedings of the 27th International Conference on Machine
  Learning (ICML-10)}
\bpages{1167--1174}.
\bpublisher{Citeseer}.
\end{binproceedings}
\endbibitem

\bibitem{xu2020non}
\begin{barticle}[author]
\bauthor{\bsnm{Xu},~\bfnm{Tengyu}\binits{T.}},
  \bauthor{\bsnm{Wang},~\bfnm{Zhe}\binits{Z.}} \AND
  \bauthor{\bsnm{Liang},~\bfnm{Yingbin}\binits{Y.}}
(\byear{2020}).
\btitle{Non-asymptotic Convergence Analysis of Two Time-scale (Natural)
  Actor-Critic Algorithms}.
\bjournal{arXiv preprint arXiv:2005.03557}.
\end{barticle}
\endbibitem

\bibitem{xu2020improving}
\begin{barticle}[author]
\bauthor{\bsnm{Xu},~\bfnm{Tengyu}\binits{T.}},
  \bauthor{\bsnm{Wang},~\bfnm{Zhe}\binits{Z.}} \AND
  \bauthor{\bsnm{Liang},~\bfnm{Yingbin}\binits{Y.}}
(\byear{2020}).
\btitle{Improving Sample Complexity Bounds for Actor-Critic Algorithms}.
\bjournal{preprint arXiv:2004.12956}.
\end{barticle}
\endbibitem

\bibitem{xu2021doubly}
\begin{barticle}[author]
\bauthor{\bsnm{Xu},~\bfnm{Tengyu}\binits{T.}},
  \bauthor{\bsnm{Yang},~\bfnm{Zhuoran}\binits{Z.}},
  \bauthor{\bsnm{Wang},~\bfnm{Zhaoran}\binits{Z.}} \AND
  \bauthor{\bsnm{Liang},~\bfnm{Yingbin}\binits{Y.}}
(\byear{2021}).
\btitle{Doubly Robust Off-Policy Actor-Critic: Convergence and Optimality}.
\bjournal{Preprint arXiv:2102.11866}.
\end{barticle}
\endbibitem

\bibitem{zhan2021policy}
\begin{barticle}[author]
\bauthor{\bsnm{Zhan},~\bfnm{Wenhao}\binits{W.}},
  \bauthor{\bsnm{Cen},~\bfnm{Shicong}\binits{S.}},
  \bauthor{\bsnm{Huang},~\bfnm{Baihe}\binits{B.}},
  \bauthor{\bsnm{Chen},~\bfnm{Yuxin}\binits{Y.}},
  \bauthor{\bsnm{Lee},~\bfnm{Jason~D}\binits{J.~D.}} \AND
  \bauthor{\bsnm{Chi},~\bfnm{Yuejie}\binits{Y.}}
(\byear{2021}).
\btitle{Policy mirror descent for regularized reinforcement learning: A
  generalized framework with linear convergence}.
\bjournal{arXiv preprint arXiv:2105.11066}.
\end{barticle}
\endbibitem

\bibitem{zhang2019convergence}
\begin{binproceedings}[author]
\bauthor{\bsnm{Zhang},~\bfnm{Kaiqing}\binits{K.}},
  \bauthor{\bsnm{Koppel},~\bfnm{Alec}\binits{A.}},
  \bauthor{\bsnm{Zhu},~\bfnm{Hao}\binits{H.}} \AND
  \bauthor{\bsnm{Ba{\c{s}}ar},~\bfnm{Tamer}\binits{T.}}
(\byear{2019}).
\btitle{Convergence and iteration complexity of policy gradient method for
  infinite-horizon reinforcement learning}.
In \bbooktitle{2019 IEEE 58th Conference on Decision and Control (CDC)}
\bpages{7415--7422}.
\bpublisher{IEEE}.
\end{binproceedings}
\endbibitem

\bibitem{zhang2021finite}
\begin{barticle}[author]
\bauthor{\bsnm{Zhang},~\bfnm{Kaiqing}\binits{K.}},
  \bauthor{\bsnm{Yang},~\bfnm{Zhuoran}\binits{Z.}},
  \bauthor{\bsnm{Liu},~\bfnm{Han}\binits{H.}},
  \bauthor{\bsnm{Zhang},~\bfnm{Tong}\binits{T.}} \AND
  \bauthor{\bsnm{Basar},~\bfnm{Tamer}\binits{T.}}
(\byear{2021}).
\btitle{Finite-sample analysis for decentralized batch multi-agent
  reinforcement learning with networked agents}.
\bjournal{IEEE Transactions on Automatic Control}.
\end{barticle}
\endbibitem

\bibitem{zhang2020provably}
\begin{binproceedings}[author]
\bauthor{\bsnm{Zhang},~\bfnm{Shangtong}\binits{S.}},
  \bauthor{\bsnm{Liu},~\bfnm{Bo}\binits{B.}},
  \bauthor{\bsnm{Yao},~\bfnm{Hengshuai}\binits{H.}} \AND
  \bauthor{\bsnm{Whiteson},~\bfnm{Shimon}\binits{S.}}
(\byear{2020}).
\btitle{Provably convergent two-timescale off-policy actor-critic with function
  approximation}.
In \bbooktitle{International Conference on Machine Learning}
\bpages{11204--11213}.
\bpublisher{PMLR}.
\end{binproceedings}
\endbibitem

\end{thebibliography}
